\documentclass[journal]{IEEEtran}

\usepackage[switch]{lineno} 

\usepackage{multirow}
\usepackage{epsfig}
\usepackage{graphicx}
\usepackage{amsmath}
\usepackage{amssymb}
\usepackage{subfigure}
\usepackage{xcolor}
\usepackage{footnote}
\usepackage{algorithm}
\usepackage{algpseudocode}
\usepackage[numbers]{natbib}
\usepackage{makecell}

\usepackage[pagebackref=false,breaklinks=true,letterpaper=true,colorlinks,bookmarks=false]{hyperref}

\graphicspath{{./figures/}{./figures/draw/}{./fig/}{./fig/bio/}}

\def\x{{\mathbf x}}

\def\w{{\mathbf w}}

\def\cc{{\mathbf c}}
\def\bb{{\mathbf b}}
\def\I{{\mathbf 1}}
\def\q{{\mathbf q}}

\def\B{{\mathbf B}}
\def\F{{\mathbf F}}
\def\X{{\mathbf X}}

\def\R{{\mathbb R}}
\def\H{{\mathbf H}}
\def\Z{{\mathbf Z}}
\def\W{{\mathbf W}}
\def\V{{\mathbf V}}
\def\Q{{\mathbf Q}}
\def\I{{\mathbf I}}
\def\1{{\mathbf 1}}
\def\S{{\mathbf S}}
\def\V{{\mathbf V}}
\newcommand\norm[1]{\left\lVert#1\right\rVert}

\newcommand\redd[1]{{\color{black}#1}} 
\newcommand\red[1]{{\color{black}#1}} 

\begin{document}
\title{Compact Hash Code Learning with \\Binary Deep Neural Network}
\author{Thanh-Toan Do, Tuan Hoang, Dang-Khoa Le Tan, Anh-Dzung Doan, Ngai-Man Cheung

	\thanks{
	Thanh-Toan Do is with the University of Liverpool, UK. Email: thanh-toan.do@liverpool.ac.uk.
	}
	\thanks{
	Tuan Hoang, Dang-Khoa Le Tan,  Ngai-Man Cheung are with Singapore University of Technology and Design, Singapore. Email: nguyenanhtuan\_hoang@mymail.sutd.edu.sg, \{letandang\_khoa, ngaiman\_cheung\}@sutd.edu.sg.
	}
	\thanks{
	Anh-Dzung Doan is with the University of Adelaide, Australia. Email: dung.doan@adelaide.edu.au.
	}

}


\maketitle

\begin{abstract}

Learning compact binary codes for image retrieval
problem using deep neural networks has recently attracted increasing
attention. However, training deep hashing networks is
challenging due to the binary constraints on the hash codes. 
In this paper, we propose deep network models and learning algorithms for learning binary hash codes given image representations under both unsupervised and supervised manners. 
The novelty of our network design is that we constrain one hidden layer to directly output the binary codes. This design has overcome a challenging problem in some previous works: optimizing non-smooth objective functions because of binarization. 
In addition, we  propose to incorporate independence and balance properties in the direct and strict forms into the learning schemes. We also include a similarity preserving property in our objective functions. 
The resulting optimizations involving these binary, independence, and balance constraints are difficult to solve. To tackle this difficulty, we propose to learn the networks with alternating optimization and careful relaxation. 
Furthermore, by leveraging the powerful capacity of convolutional neural networks, we propose an end-to-end architecture that jointly learns to extract visual features and produce binary hash codes. 
Experimental results for the benchmark datasets show that the proposed methods compare favorably or outperform the state of the art.

\end{abstract}
\section{Introduction}
Content-based image retrieval is an important and well studied problem in computer vision. It has many applications such as the visual search ~\cite{DBLP:conf/cvpr/ArandjelovicZ13,herve_cvpr2010,do2017embedding,DBLP:conf/cvpr/JegouZ14,do_cvpr15}, place recognition~\cite{247,netvlad,Torii-PAMI2015}, and camera pose estimation~\cite{Sattler1,Irschara,man-tip19}, to name a few. 
In the state-of-the-art image retrieval systems~\cite{DBLP:conf/cvpr/ArandjelovicZ13,herve_cvpr2010,do2017embedding,DBLP:conf/cvpr/JegouZ14,do_cvpr15,selectiveConv}, images are represented as high-dimensional feature vectors that can later be searched via the classical distance such as the Euclidean or Cosine distance. 
However, when the database is scaled up, there are two main requirements for retrieval systems, i.e., efficient storage and fast searching. Among solutions, binary hashing is an attractive approach for achieving those requirements~\cite{Grauman_review,survey_Hash4simSearch,survey_learn2hash,DBLP:journals/corr/WangLKC15,lsh_vldb09,KLSH_iccv09,KLSH_nips09,DBLP:journals/pami/KulisJG09,SpH,ITQ,kmeanHash,SpH_CVPR12,conf/nips/KongL12,DBLP:journals/tmm/ZhuLZXYW17,DBLP:journals/tmm/WangCO015,DBLP:journals/tmm/MaoYL17,LDAHash,SupHashKernel,DBLP:journals/pami/WangKC12,DBLP:conf/nips/0002FS12,DBLP:conf/iccv/LinSSH13,BRE,SDH_CVPR15,do2016-sdp,toan-tip19,Tuan:tom:2019}. 
Briefly, binary hashing learns a mapping (hashing) function that maps each original high dimensional vector $\x \in \R^D$ into a very compact binary vector $\bb \in \{-1,1\}^L$, where $L \ll D$. 
The distances between binary data points can be efficiently calculated by the bit operations, i.e., XOR
and POPCOUNT. Furthermore, the binary representations also
result in the sufficient storage. 

The hashing methods can be divided into two groups, i.e.,  data-independent and data-dependent methods. The former ones~\cite{lsh_vldb09,KLSH_iccv09,KLSH_nips09,DBLP:journals/pami/KulisJG09} rely on random projections to construct hash functions. The representative methods
in this category are Locality Sensitive Hashing (LSH)~\cite{lsh_vldb09} and
its kernelized or discriminative extensions~\cite{KLSH_iccv09,KLSH_nips09}. 
The latter ones use the available training data to learn the hash functions in unsupervised 
\cite{SpH,ITQ,kmeanHash,SpH_CVPR12,conf/nips/KongL12,DBLP:journals/tmm/ZhuLZXYW17,DBLP:journals/tmm/WangCO015,DBLP:journals/tmm/MaoYL17,do2016-sdp,toan-tip19}  
or (semi-)supervised 
\cite{LDAHash,SupHashKernel,DBLP:journals/pami/WangKC12,DBLP:conf/nips/0002FS12,DBLP:conf/iccv/LinSSH13,BRE,SDH_CVPR15} manners. 
The representative unsupervised hashing methods, such as Spherical Hashing~\cite{SpH_CVPR12}, Spectral Hashing~\cite{SpH}, Iterative Quantization (ITQ)~\cite{ITQ}, and K-means Hashing~\cite{kmeanHash}, attempt to learn binary codes that preserve similar neighbors and the local structure of samples. 
The representative supervised hashing methods, such as, ITQ-CCA~\cite{ITQ}, Binary Reconstructive Embedding \cite{BRE}, Kernel Supervised Hashing~\cite{SupHashKernel}, Two-step Hashing~\cite{DBLP:conf/iccv/LinSSH13}, and Supervised Discrete Hashing~\cite{SDH_CVPR15}, attempt to learn binary codes that preserve the label similarity between samples.  
Extensive reviews of data-independent and  data-dependent hashing methods can be found in recent surveys~\cite{Grauman_review,survey_Hash4simSearch,survey_learn2hash,DBLP:journals/corr/WangLKC15}.


One problem that makes the binary hashing difficult is the binary constraint on the codes, i.e., the outputs of the hash functions must be binary. Generally, this binary constraint leads to an NP-hard mixed-integer optimization problem. 
To overcome this difficulty, most of the aforementioned methods have relied on relaxation approaches that relax the constraint during the learning of hash functions. 
By using the relaxation approach, the continuous codes are first learned. Subsequently, the codes are binarized, for example, by thresholding. The relaxation significantly simplifies the original constrained binary problem.  However,  the solution can be suboptimal, i.e., the binary codes resulting from thresholded continuous codes could be inferior to those obtained by directly including the binary constraint in the learning.

In addition, as shown in the notable work \textit{Spectral Hashing}~\cite{SpH}, good binary codes should also have the following properties: (i) similarity preservation -- (dis)similar inputs should likely have (dis)similar binary codes; (ii) independence -- different bits in the binary codes are independent of each other so that no redundant information is captured; (iii) bit balance --  each bit has a $50\%$ chance of being $1$ or $-1$.
It is worth noting that direct incorporation of the independent and balance properties can complicate the learning. Previous works~\cite{ITQ, DBLP:journals/pami/WangKC12} have used some relaxation or approximation to overcome the difficulties, but there may be some performance degradation.

Recently, deep learning has attracted great attention from the computer vision community due to its superiority in many vision tasks such as classification, detection, and segmentation~\cite{DBLP:conf/nips/KrizhevskySH12,DBLP:conf/cvpr/RazavianASC14,DBLP:conf/cvpr/LongSD15}. Inspired by the success of deep learning in different vision tasks, recently, some researchers have used  deep learning 
for joint learning image representations and binary hash codes in an end-to-end deep learning-based supervised hashing framework~\cite{DSRH,DRSCH,simulfeature,DQN}. However, learning binary codes in deep networks is challenging. This is because one has to deal with the binary constraint on the hash codes,
i.e., one layer of the network should output binary codes. A naive solution is to adopt
the $sign$ activation layer to produce binary codes. However, due to the lack of smoothness of the $sign$ function, it causes the \textit{vanishing gradient} problem when training the network with standard back propagation~\cite{DBLP:journals/neco/HintonOT06}. 


\textbf{Contributions: } In this paper, first, we propose a novel deep network model and a learning algorithm for unsupervised hashing. Specifically, when learning binary codes, instead of involving the $sign$ or step function as in recent works~\cite{DeepHash_CVPR15, BA_CVPR15}, our proposed network design constrains one layer to directly output the binary codes (therefore, the network is named as \textit{Binary Deep Neural Network}). 
In addition, we propose to directly integrate the independence and balance properties into the objective function.
Furthermore, we also include the similarity preserving property in our objective function. 
The resulting optimization with these binary and direct constraints is NP-hard.  
To overcome this challenge, we propose to attack the problem with alternating optimization and careful relaxation. 
Second, to increase the discriminative power of the binary codes, we extend our method to supervised hashing by leveraging the label information so that the binary codes preserve the semantic similarity between samples. 
Finally, to demonstrate the flexibility of our proposed method and to leverage the power of convolutional deep neural networks, we adapt our optimization strategy and the proposed supervised hashing model to an end-to-end deep hashing  framework. Solid experiments on various benchmark datasets show the improvements of the proposed methods over state-of-the-art hashing methods. 

A preliminary version of this work has been reported in~\cite{do2016binary}. In this work, we present a substantial extension to our previous work. In particular, the main extension is that we propose the end-to-end binary deep neural network framework which jointly learns the image features and binary codes. The experimental results show that the proposed end-to-end hashing framework significantly boosts the retrieval accuracy. Other extensions are more extensive experiments (e.g., new experiments on the SUN397 dataset~\cite{sun397}, comparison to recent state-of-the-art end-to-end unsupervised and supervised hashing methods) to evaluate the effectiveness of the proposed methods. 

The remainder of this paper is organized as follows. Section~\ref{sec:relatedwork} presents related works. Section~\ref{sec:UH-BDNN} presents and evaluates the proposed unsupervised hashing method. Section~\ref{sec:SH-BDNN} presents and evaluates the proposed supervised hashing method. Section~\ref{sec:SH-E2E-BDNN} presents and evaluates the proposed end-to-end deep hashing network. Section~\ref{sec:conclusion} concludes the paper. 
\section{Related work} 
\label{sec:relatedwork}
In this section, we review related works that have also used neural networks as hash functions.
We review both works that have used shallow network architectures~\cite{SemanticH, DeepHash_CVPR15, DeepHash_TIP17, BA_CVPR15} and recent works which use end-to-end deep architectures~\cite{deepbit2016,DSRH,DRSCH,simulfeature,DHN,hashnet,DPSH}. 

In Semantic Hashing~\cite{SemanticH}, the authors design a deep model by stacking Restricted Boltzmann Machines. 
That model does not consider the independence and balance of the codes. 
In Binary Autoencoder~\cite{BA_CVPR15}, the hash function is defined as a linear autoencoder. 
Because the Binary Autoencoder model~\cite{BA_CVPR15} only uses one hidden layer, it may not well capture the input informations.
We note that extension~\cite{BA_CVPR15} with multiple, nonlinear layers is not easy due to the binary constraint. The model in \cite{BA_CVPR15} also does not consider the independence and balance of codes. In Deep Hashing~\cite{DeepHash_CVPR15,DeepHash_TIP17},  the hash function is defined as a deep neural network. Nevertheless, the Deep Hashing model does not fully take into account the similarity preserving property. 
Furthermore, the authors also apply some relaxations in arriving at the independence and balance of codes. Those relaxations may degrade the performance. 
\redd{Recently, several works \cite{deepbit2016,ijcai2018-148,8247210} have leveraged the Convolutional Neural Network (CNN) to learn more discriminative hash codes in the unsupervised manner. 
In DeepBit \cite{deepbit2016}, the softmax layer of a pretrained network (i.e., VGG~\cite{VGG}) is replaced by a hash layer. Its loss function enforces several criteria on the codes produced by the hash layer, i.e., the output codes should: minimize the quantization loss; be distributed evenly; be invariant to rotation. The authors assume that the fully connected features produced by the pre-trained network are already sufficiently discriminative for image retrieval task. Hence, no similarity preserving criterion is considered on the hash codes. 
In Similarity-Adaptive Deep Hashing (SADH) \cite{8247210}, the authors propose to alternatively proceed over three training modules: deep hash model training, similarity graph updating and binary code optimization (with graph hashing \cite{SpH}).
In \cite{ijcai2018-148}, the authors propose to analyze semantic informative deep features to obtain a semantic similarity matrix $\S$. The authors also relax the $sign$ function to the $tanh$ function to avoid the ill-posed gradient problem.}

To handle the binary constraint, in Semantic Hashing~\cite{SemanticH}, the authors first solves the relaxed problem by ignoring the binary constraints. Then, they apply thresholding on the continuous solution which results in binary codes. In Deep Hashing (DH)~\cite{DeepHash_CVPR15,DeepHash_TIP17}, the binary codes are achieved by applying the $sign$ function on the outputs of the last layer, $\H^{(n)}$.  The authors include a term in the objective function to reduce this binarization loss: $\left(sign(\H^{(n)}) - \H^{(n)} \right)$. However, due to the non-differentiability of the $sign$ function,  
solving the objective function of DH~\cite{DeepHash_CVPR15,DeepHash_TIP17} is difficult. 
The authors in~\cite{DeepHash_CVPR15,DeepHash_TIP17} assumed that the $sign$ function is differentiable everywhere, i.e., the derivative of $sign(x)$ equals zero for all values of $x$. 
In Binary Autoencoder (BA)~\cite{BA_CVPR15},  the binary codes are achieved by passing the outputs of the hidden layer into a step function.  Incorporating the step function in the learning leads to a non-smooth objective function, i.e., a NP-complete problem. To handle this challenge, the authors \cite{BA_CVPR15} use binary SVMs to learn the model parameters in the case when there is only a single hidden layer.

Joint learning image representations and binary hash codes in an end-to-end deep learning-based supervised hashing framework~\cite{DSRH,DRSCH,simulfeature,DBLP:conf/mm/ZhangWHC16,hashnet,DSH,DHN,DPSH} have shown a considerable boost in retrieval accuracy. By joint optimization, the produced hash codes are more sufficient to preserve the semantic similarity between images. In those works, the network architectures often consist of a feature extraction sub-network and a subsequent hashing layer to produce hash codes. 
Ideally, the hashing layer should adopt a $sign$ activation function to output exactly binary codes. However, due to the vanishing gradient difficulty of the $sign$ function, an approximation procedure must be employed. For example, $sign$ can be approximated by a tanh-like function $y=tanh(\beta x)$, where $\beta$ is a free parameter controlling the trade off between the smoothness and the binary quantization loss \cite{DRSCH}. However, it is non-trivial to determine an optimal $\beta$. A small $\beta$ causes large binary quantization loss while a large $\beta$ makes the output of the function close to the binary values, but the gradient of the function almost vanishes, making back-propagation infeasible. 
\red{In \cite{DRSCH}, the $\beta$ value is heuristically increased gradually reducing the smoothness as the training proceeds.}
\red{Recently, similar to \cite{DRSCH}, HashNet \cite{hashnet} handles the non-smooth problem of the $sign$ function by continuation, i.e., starting the training with a smoothed objective function and gradually reducing the smoothness as the training proceeds. 
Furthermore, the above methods do not consider the independence and balance properties of the binary codes.}
The trade off problem between the smoothness and quatization loss persists when the logistic-like functions \cite{DSRH,simulfeature} are used. 
\red{In recent deep hashing works~\cite{DSH,DHN}, the absolute function and $l_1$ regularization are used to deal with the binary constraint on the codes. However, both absolute function and $l_1$ regularization are non-differentiable. The authors work around this difficulty by assuming that both are differentiable everywhere, but there may be some performance degradation.}
\red{In Deep Pairwise-Supervised Hashing (DPSH) \cite{DPSH}, the authors  design a method to handle the binary constraint of the pairwise-supervised hashing objective function. Specifically, the outputs of the model are first computed, and then the corresponding binary codes are obtained by applying the $sign$ function to the outputs. By assuming the binary codes are fixed (to avoid the ill-posed gradient problem of the $sign$ function), the gradients are then computed to update the model weights.
DPSH also does not consider the independence and balance properties, which are important for the hashing problem \cite{SpH}. Recently, in \cite{DBLP:conf/mm/ZhangWHC16} the authors proposed an binary encoder-decoder Recurrent Neural Network for video hashing. To handle the non-smooth problem of the \textit{sign} function, the authors proposed to
use the hinge loss to approximate the
\textit{sign} function. As will be discussed, our work proposes different formulations and new learning algorithms to deal with the binary constraints on the codes.}

\section{Unsupervised Hashing with Binary Deep Neural Network (UH-BDNN)}
\label{sec:UH-BDNN}
\subsection{Formulation of UH-BDNN}
\label{subsec:formular_un}

\begin{table}[!t]
\footnotesize
\caption{Notations and their corresponding meanings.}
\centering
\begin{tabular}{|l|l|}
\hline
Notation	&Meaning \\ \hline
$\X$ &$\X = \{\x_i\}_{i=1}^{m} \in \R^{D\times m}$: set of $m$ training samples; \\
	 &each column of $\X$ corresponds to one sample\\ \hline	
$\B$ &$\B = \{\bb_i\}_{i=1}^{m} \in \{-1,+1\}^{L\times m}$: binary code of $\X$ \\ \hline
$L$  &Number of required bits to encode a sample \\ \hline
$n$  &Number of layers (including input and output layers) \\ \hline
$s_l$&Number of units in layer $l$	\\ \hline
$f^{(l)}$ &Activation function of layer $l$ \\ \hline		
$\W^{(l)}$&$\W^{(l)} \in \R^{s_{l+1}\times s_l}$: weight matrix connecting layer $l+1$ 	\\ 
		&  and layer $l$ \\ \hline
$\cc^{(l)}$&$\cc^{(l)} \in \R^{s_{l+1}}$:bias vector for units in layer $l+1$ \\ \hline
$\H^{(l)}$ &$\H^{(l)} = f^{(l)}\left(\W^{(l-1)}\H^{(l-1)} + \cc^{(l-1)}\1_{1\times m}\right)$:  \\ 
	      &output values of layer $l$; convention: $\H^{(1)} = \X$ \\ \hline
$\1_{a\times b}$ & Matrix has $a$ rows, $b$ columns and all elements equal to 1	\\ \hline
\end{tabular}
\label{tab:notation}
\end{table}


\begin{figure}[!t]
\centering
\includegraphics[scale=0.35]{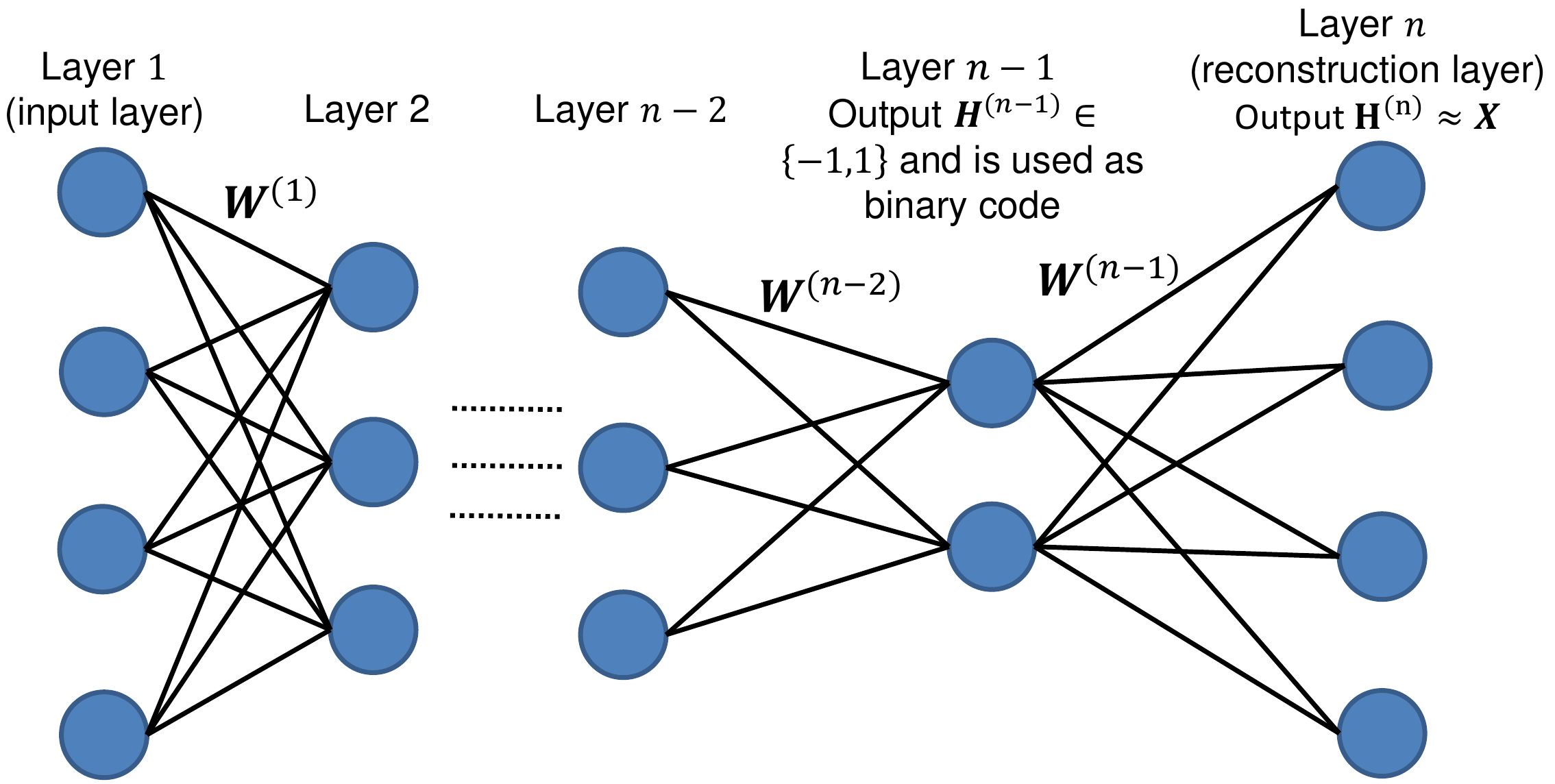}
\caption{The illustration of our UH-BDNN ($D=4,L=2$). In our proposed network design, the outputs of layer $n-1$ are constrained to $\{-1,1\}$ and are used as the binary codes. During training, these codes are used to reconstruct the input samples at the final layer.}
\label{fig:network}
\end{figure}
For easy following, we first summarize the notations in Table~\ref{tab:notation}.  In our work, the hash functions are defined by a deep neural network. In our proposed architecture, we use different activation functions in different layers.  Specifically, we use the sigmoid function as the activation function for layers $2,\cdots,n-2$, and the identity function as the activation function for layer $n-1$ and layer $n$. 
Our idea is to learn the network such that the output values of the \textit{penultimate layer} (layer $n-1$)
can be used as the binary codes.  We introduce constraints in the learning algorithm such that the output values at the layer $n-1$ 
have the following desirable properties: \textit{(i)} belonging to $\{-1,1\}$; \textit{(ii)} similarity preserving; \textit{(iii)} independence and \textit{(iv)} balancing. Fig.~\ref{fig:network} illustrates our  network for the case $D=4,L=2$.

Let us start with the first two properties of the codes, i.e., belonging to $\{-1,1\}$ and similarity preserving. To achieve binary codes having these two properties, we propose to optimize the following constrained objective function

\vspace{-0.3em} 
\footnotesize
\begin{eqnarray}
\min_{\W,\cc} J &=& \frac{1}{2m} \norm{\X-\left(\W^{(n-1)}\H^{(n-1)}+\cc^{(n-1)}\1_{1\times m}\right)}^2 \nonumber \\
{} && +\frac{\lambda_1}{2}\sum_{l=1}^{n-1} \norm{\W^{(l)}}^2
 \label{eq:obj_ori}
\end{eqnarray}
\begin{equation}
\hspace{-2cm}\textrm{s.t. } \H^{(n-1)} \in \{-1,1\}^{L\times m} \label{eq:binary0}
\end{equation} 
\normalsize 
Constraint~(\ref{eq:binary0}) is to ensure the first property. As the activation function for the last layer is the identity function, the term $\left(\W^{(n-1)}\H^{(n-1)}+\cc^{(n-1)}\1_{1\times m}\right)$ is the output of the last layer. The first term of~(\ref{eq:obj_ori}) ensures that the binary code gives a good reconstruction of $\X$. 
It is worth noting that the reconstruction criterion has been used as an indirect approach for preserving the similarity in state-of-the-art unsupervised hashing methods~\cite{ITQ, BA_CVPR15,SemanticH}, i.e., it encourages (dis)similar inputs to be mapped to (dis)similar binary codes. 
The second term is a regularization that tends to decrease the magnitude of the weights, which helps to prevent overfitting. 
Note that in our proposed design, we constrain the network to directly output the binary codes at one layer, which avoids the difficulty of the $sign$ / $step$ function which is non-differentiability. Our formulation with (\ref{eq:obj_ori}) under the binary constraint~(\ref{eq:binary0}) is difficult to solve. It is a mixed-integer problem that is NP-hard. To address the problem, we propose to introduce an auxiliary variable $\B$ and use alternating optimization. Consequently, we reformulate the objective function~(\ref{eq:obj_ori}) under constraint~(\ref{eq:binary0}) as follows:          

\vspace{-0.3em} 
\footnotesize 
\begin{eqnarray}
\min_{\W,\cc,\B} J &=& \frac{1}{2m} \norm{\X-\W^{(n-1)}\B-\cc^{(n-1)}\1_{1\times m}}^2 \nonumber \\
{} && +\frac{\lambda_1}{2}\sum_{l=1}^{n-1} \norm{\W^{(l)}}^2
 \label{eq:obj_2}
\end{eqnarray}
\begin{equation}
\textrm{s.t. }\B = \H^{(n-1)}, \label{eq:binary}
\end{equation} 
\vspace{-0.3cm}
\begin{equation}
\hspace{3.7em}\B \in \{-1,1\}^{L\times m}. \label{eq:binary_add}
\end{equation}
\normalsize

The advantage of introducing the auxiliary variable $\B$ is that the difficult constrained optimization problem~(\ref{eq:obj_ori}) can be decomposed into two simpler sub-optimization problems. Consequently, we are able to iteratively solve the optimization problem by using alternating optimization with respect to $(\W,\cc)$ and $\B$ while holding the other fixed. 
Inspired from the quadratic penalty method~\cite{Nocedal06}, we relax the equality constraint (\ref{eq:binary}) by converting it into a penalty term. We achieve the following constrained objective function 

\vspace{-0.3em}\footnotesize
\begin{eqnarray}
\min_{\W,\cc,\B} J &=& \frac{1}{2m} \norm{\X-\W^{(n-1)}\B-\cc^{(n-1)}\1_{1\times m}}^2 \nonumber \\ 
{}&&\hspace{0em}+\frac{\lambda_1}{2}\sum_{l=1}^{n-1} \norm{\W^{(l)}}^2 + \frac{\lambda_2}{2m}\norm{\H^{(n-1)}-\B}^2  \label{eq:obj_4}
\end{eqnarray}
\begin{equation}
\hspace{-1.5cm}\textrm{s.t. }\B \in \{-1,1\}^{L\times m} \label{eq:binary_H}
\end{equation}
\normalsize
in which, the third term in (\ref{eq:obj_4}) measures the (equality) constraint violation. By setting the penalty parameter $\lambda_2$ sufficiently large, we penalize the constraint violation severely, thereby forcing the minimizer of the penalty function (\ref{eq:obj_4}) closer to the feasible region of the original constrained function (\ref{eq:obj_2}). 

Now let us consider the independence and balance properties of the codes.  
We note that the independence and balance can be constrained in $\B$. However, this makes the optimization on $\B$  difficult. Thus, for independence and balance, we constrain on $\H^{(n-1)}$. 
In contrast to previous works that use some relaxation or approximation on the independence and balance properties~\cite{ITQ, DeepHash_CVPR15,DBLP:journals/pami/WangKC12}, in this work, we propose to encode these properties strictly and directly based on the outputs of the layer $n-1$.  
 In particular, we encode the independence and balance properties of the codes by introducing the fourth and the fifth terms respectively in the following constrained objective function

\vspace{-0.2cm}
\footnotesize
\begin{eqnarray}
\min_{\W,\cc,\B} J &=& \frac{1}{2m} \norm{\X-\W^{(n-1)}\B-\cc^{(n-1)}\1_{1\times m}}^2 \nonumber \\ 
{}&&\hspace{-5em} +\frac{\lambda_1}{2}\sum_{l=1}^{n-1} \norm{\W^{(l)}}^2 + \frac{\lambda_2}{2m}\norm{\H^{(n-1)}-\B}^2 \nonumber \\
{}&&\hspace{-5em}+\frac{\lambda_3}{2}\norm{\frac{1}{m}\H^{(n-1)}(\H^{(n-1)})^T-\I}^2 +\frac{\lambda_4}{2m}\norm{\H^{(n-1)}\1_{m\times 1}}^2 \label{eq:obj_5}
\end{eqnarray}
\begin{equation}
\hspace{-2cm}\textrm{s.t. }\B \in \{-1,1\}^{L\times m}. \label{eq:binary_H1}
\end{equation}
\normalsize
The objective function (\ref{eq:obj_5}) under constraint (\ref{eq:binary_H1}) is our final formulation.

\subsection{Optimization}
\label{subsec:UDH_opt}
To solve~(\ref{eq:obj_5}) under constraint~(\ref{eq:binary_H1}), we propose to use  alternating optimization over $(\W,\cc)$ and $\B$.

\subsubsection{$(\W,\cc)$ step}
\label{subsub:W_step_un}
When fixing $\B$, the problem becomes the unconstrained optimization. This is solved by using the \textit{L-BFGS}~\cite{L-BFGS} optimizer with backpropagation. The gradients of the objective function $J$ (\ref{eq:obj_5}) w.r.t. different parameters are computed as follows.

At $l = n-1$, we have

\vspace{-0.2cm}
\footnotesize
\begin{equation}
\frac{\partial J}{\partial \W^{(n-1)}} = \frac{-1}{m}(\X-\W^{(n-1)}\B-\cc^{(n-1)}\1_{1\times m})\B^T + \lambda_1 \W^{(n-1)}
\end{equation}
\begin{equation}
\frac{\partial J}{\partial \cc^{(n-1)}} = \frac{-1}{m}\left( (\X-\W^{(n-1)}\B)\1_{m\times 1}-m\cc^{(n-1)} \right)
\end{equation}
\normalsize
For other layers, let us define

\vspace{-0.2cm}
\footnotesize
\begin{eqnarray}
\Delta^{(n-1)} &=& \left[ \frac{\lambda_2}{m}\left( \H^{(n-1)}-\B \right)\right. \nonumber \\
{}&& \left. +\frac{2\lambda_3}{m}\left( \frac{1}{m}\H^{(n-1)}(\H^{(n-1)})^T - \I\right)\H^{(n-1)} \right. \nonumber \\
 {}&& \left. +\frac{\lambda_4}{m}\left( \H^{(n-1)}\1_{m\times m} \right) \right]\odot f^{(n-1)'}(\Z^{(n-1)})
\end{eqnarray}
\begin{equation}
\Delta^{(l)} = \left( (\W^{(l)})^T\Delta^{(l+1)} \right) \odot f^{(l)'}(\Z^{(l)}),\forall l = n-2,\cdots,2
\end{equation}
\normalsize
where \small$\Z^{(l)} = \W^{(l-1)}\H^{(l-1)} + \cc^{(l-1)}\1_{1\times m}$, $l=2,\cdots,n$; $\odot$ denotes the Hadamard product. \normalsize

Then, $\forall l = n-2,\cdots,1$, we have

\footnotesize
\begin{equation}
\frac{\partial J}{\partial \W^{(l)}} = \Delta^{(l+1)}(\H^{(l)})^T +\lambda_1\W^{(l)}
\end{equation}
\begin{equation}
\frac{\partial J}{\partial \cc^{(l)}} = \Delta^{(l+1)}\1_{m\times 1}
\end{equation}

\normalsize
\subsubsection{$\B$ step}
\label{subsub:B_step_un}
When fixing $(\W,\cc)$, we can rewrite problem~(\ref{eq:obj_5}) as 

\vspace{-0.5em}
\footnotesize
\begin{eqnarray}
\min_{\B} J &=& \norm{\X-\W^{(n-1)}\B-\cc^{(n-1)}\1_{1\times m}}^2 \nonumber \\
{}&& +\lambda_2 \norm{\H^{(n-1)}-\B}^2
\label{eq:B_step_un}
\end{eqnarray} 
\begin{equation}
\hspace{-1cm}\textrm{s.t. }\B \in \{-1,1\}^{L\times m}. \label{eq:binary_H11}
\end{equation}
\normalsize
We adaptively use the recent method \textit{discrete cyclic coordinate descent}~\cite{SDH_CVPR15} to  iteratively solve  $\B$, i.e., row by row. The advantage of this method is that if we fix $L-1$ rows of $\B$ and only solve for the remaining row, we can achieve the closed-form solution for that row. 

\smallskip
Let $\V = \X-\cc^{(n-1)}\1_{1\times m}$; $\Q = (\W^{(n-1)})^T\V+\lambda_2\H^{(n-1)}$. For $k=1,\cdots L$, let $\w_k$ be $k^{th}$ column of $\W^{(n-1)}$; $\W_1$ be matrix $\W^{(n-1)}$ excluding $\w_k$; $\q_k$ be $k^{th}$ column of $\Q^T$; $\bb_k^T$ be $k^{th}$ row of $\B$; $\B_1$ be the matrix of $\B$ excluding $\bb_k^T$. We have the closed-form for $\bb_k^T$ as
\footnotesize
\begin{equation}
\bb_k^T = sign(\q^T - \w_k^T\W_1\B_1).
\end{equation}
\normalsize
The proposed UH-BDNN method is summarized in Algorithm $\mathbf{1}$. In Algorithm $\mathbf{1}$, $\B_{(t)}$ and $(\W,\cc)_{(t)}$ are values of $\B$ and $\{\W^{(l)},\cc^{(l)}\}_{l=1}^{n-1}$ at iteration $t$, respectively.

\begin{algorithm}[!t]
    \footnotesize
	\caption{Unsupervised Hashing with Binary Deep Neural Network (UH-BDNN)}
	\begin{algorithmic}[1] 
		\Require 
			\Statex $\X = \{\x_i\}_{i=1}^{m} \in \R^{D\times m}$: training data; $L$: code length; $T$: maximum iteration number; $n$: number of layers; $\{s_l\}_{l=2}^{n}$: number of units of layers $2 \to n$ (note: $s_{n-1} = L$, $s_n = D$); $\lambda_1, \lambda_2, \lambda_3, \lambda_4$.
		\Ensure 
			\Statex 
			Parameters $\{\W^{(l)},\cc^{(l)}\}_{l=1}^{n-1}$
			\Statex 
			\State Initialize $\B_{(0)} \in \{-1,1\}^{L\times m}$ using ITQ~\cite{ITQ}
			\State Initialize $\{\cc^{(l)}\}_{l=1}^{n-1} = \mathbf{0}_{s_{l+1}\times 1}$. Initialize $\{\W^{(l)}\}_{l=1}^{n-2}$ by getting the top $s_{l+1}$ eigenvectors from the covariance matrix of $\H^{(l)}$. Initialize $\W^{(n-1)} = \I_{D\times L}$
			\State Fix $\B_{(0)}$, compute $(\W,\cc)_{(0)}$ with $(\W,\cc)$ step using initialized $\{\W^{(l)},\cc^{(l)}\}_{l=1}^{n-1}$ (line 2) as starting point for L-BFGS.
			\For{$t = 1 \to T$}
				\State Fix $(\W,\cc)_{(t-1)}$, compute $\B_{(t)}$ with $\B$ step
				\State Fix $\B_{(t)}$, compute $(\W,\cc)_{(t)}$ with $(\W, \cc)$ step using $(\W,\cc)_{(t-1)}$ as starting point for L-BFGS.
			\EndFor
			\State Return 
			$(\W,\cc)_{(T)}$
    \end{algorithmic}
    \label{alg1}
\end{algorithm}



\subsection{Evaluation of Unsupervised Hashing with Binary Deep Neural Network (UH-BDNN)}
\label{sec:eva_UH-BDNN}

In this section, we conduct experiments to evaluate the effectiveness of the proposed method UH-BDNN. We compare UH-BDNN with other unsupervised hashing methods: Binary Autoencoder (BA)~\cite{BA_CVPR15}, Spectral Hashing (SH)~\cite{SpH},  Iterative Quantization (ITQ)~\cite{ITQ},  Spherical Hashing (SPH)~\cite{SpH_CVPR12}, and K-means Hashing (KMH)~\cite{kmeanHash}, which are the state-of-the-art unsupervised hashing methods. For all compared methods, we use the released implementations and the suggested parameters provided by the authors.


\subsubsection{Dataset, evaluation protocol, and implementation notes}
\label{subsec:data-imp-eva}
We evaluate and compare methods on three  benchmarking datasets: CIFAR10~\cite{cifar10}, 
MNIST~\cite{mnistlecun}, and SIFT1M~\cite{PQ}. 
\paragraph{Dataset}

The {MNIST~\cite{mnistlecun}} dataset contains 70,000 handwritten digit
images of 10 classes. We use the original split of the dataset. The training set (also used as the database for retrieval) consists of 60,000 images. The query set consists of 10,000 images. Each image is represented by a 784 dimensional feature vector by using its intensity in gray-scale.

The {CIFAR10~\cite{cifar10}} dataset contains 60,000 images of 10 classes. 
We use the original split of the dataset. The provided test set of 10,000 images is used as the query set. The remaining 50,000 images are used as the training set and the database for the retrieval. 
Each image is represented by an 800-dimensional feature vector extracted by PCA from the 4096-dimensional CNN feature produced by AlexNet~\cite{DBLP:conf/nips/KrizhevskySH12}.

The {SIFT1M~\cite{PQ}} dataset is used to evaluate the proposed method on a large scale. The dataset contains 128 dimensional SIFT vectors~\cite{SIFT_Lowe}. The original split of this dataset consists of three separated sets of 1M, 100K and 10K vectors. These three sets correspond to the database, training, and query set respectively.

\paragraph{Evaluation protocol} 
In the state-of-the-art unsupervised hashing~\cite{ITQ,SpH_CVPR12,kmeanHash, BA_CVPR15}, during the evaluation, instead of using the class labels, the Euclidean nearest neighbors are used as the  ground truths for the queries. Hence, we follow this setting to evaluate the proposed method. 
Specifically, the number of ground truths are set as in~\cite{BA_CVPR15}. For each query image of the CIFAR10 and MNIST datasets, we use its 50 Euclidean nearest neighbors as the ground truths. For the large scale dataset SIFT1M, we use 10,000 Euclidean nearest neighbors as the ground truths for each query. 
To evaluate the retrieval performance, we follow the state of the art~\cite{ITQ, BA_CVPR15, DeepHash_CVPR15} which use the following evaluation metrics: 1) mean Average Precision (mAP); 2) precision of Hamming radius $2$ (precision$@2$) which measures precision on retrieved images with a Hamming distance to query $\le 2$ (note that we report zero precision in the case of no satisfactory image). Because computing mAP is slow for the large dataset SIFT1M, we consider the top 10,000 returned neighbors when computing mAP.

\paragraph{Implementation notes}
In our deep model, we use $n=5$ layers. More specifically, for the code lengths of 8, 16, 24 and 32 bits, the numbers of units in hidden layers $2,3$, and $4$ are empirically set as $[90 \to 20 \to 8]$, $[90 \to 30 \to 16]$, $[100 \to 40 \to 24]$ and $[120 \to 50 \to 32]$ respectively. 
Additionally, the parameters $\lambda_1$, $\lambda_2$, $\lambda_3$ and $\lambda_4$ are empirically set by cross validation 
as $10^{-5}$, $5\times 10^{-2}$, $10^{-2}$ and $10^{-6}$, respectively. Finally, we empirically set the max iteration number $T$ to 10. 


\subsubsection{Retrieval results}

\begin{table}[t]
\centering
\caption{The effects of our proposed independence (IND) and balance (BAL) terms on retrieval performances (mAP). The experiments are conducted on CIFAR10 and MNIST datasets.}
\label{tb:term_effect}
\scalebox{0.96}{
\begin{tabular}{|l|c|c|c|c|c|c|}
\hline
 & \multicolumn{3}{c|}{CIFAR10} &  \multicolumn{3}{c|}{MNIST}\\ \hline
Obj. function & 16 & 24 & 32 & 16 & 24 & 32 \\ \hline
With IDN+BAL &  \textbf{5.80} & \textbf{9.11} & \textbf{11.97} & \textbf{10.57} & \textbf{18.32} & \textbf{24.90} \\ \hline
No BAL & 5.44 & 9.02 & 11.60 & 10.42 & 17.99 & 24.84\\ \hline
No IND & 5.13 & 8.89 & 11.28 & 9.91 & 17.85 & 24.75\\ \hline
No IND+BAL & 4.89 & 8.57 & 10.93 & 9.81 & 17.55 & 24.57 \\ \hline
\end{tabular}}
\end{table}

\redd{\paragraph{Effects of independence and balance terms}
First, we investigate the contributions of independence and balance terms in our proposed method. 
The quantitative results shown in Table \ref{tb:term_effect} clearly confirm the importance of independence and balance terms. More specifically, when including the proposed independence and balance terms, we achieve improvement on the retrieval performance (mAP) (i.e., $>0.5\%$ in the majority of experiments). 
Additionally, the experimental results also show that the independence term plays a more important role than the balance term, i.e., the performance drops are larger without the independence term than those without the balance term.}

\begin{figure*}[!t]
\centering
\subfigure[CIFAR10]{
\includegraphics[scale=0.32]{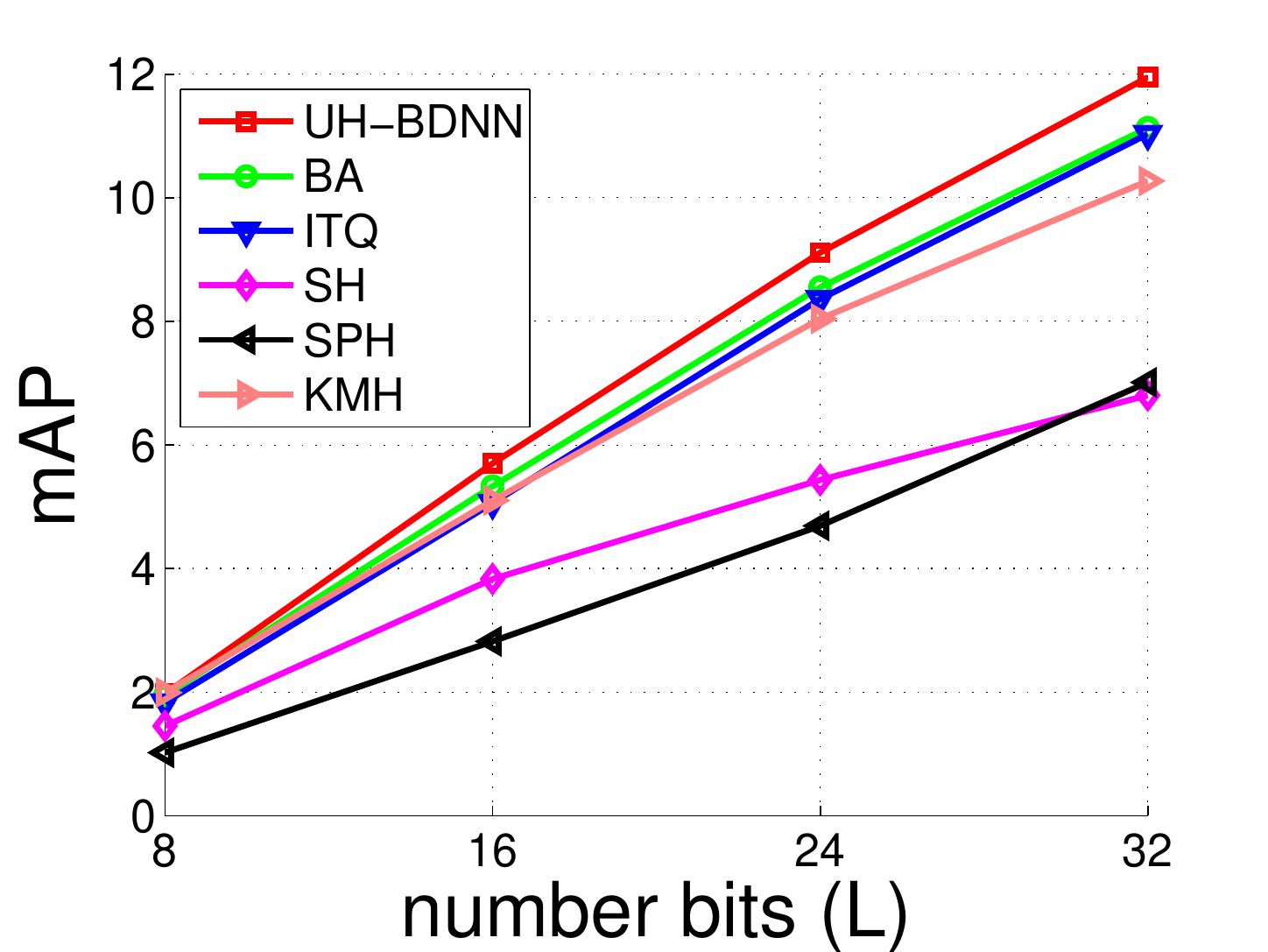}
\label{fig:cifar_mAP}
}
\subfigure[MNIST]{
\includegraphics[scale=0.32]{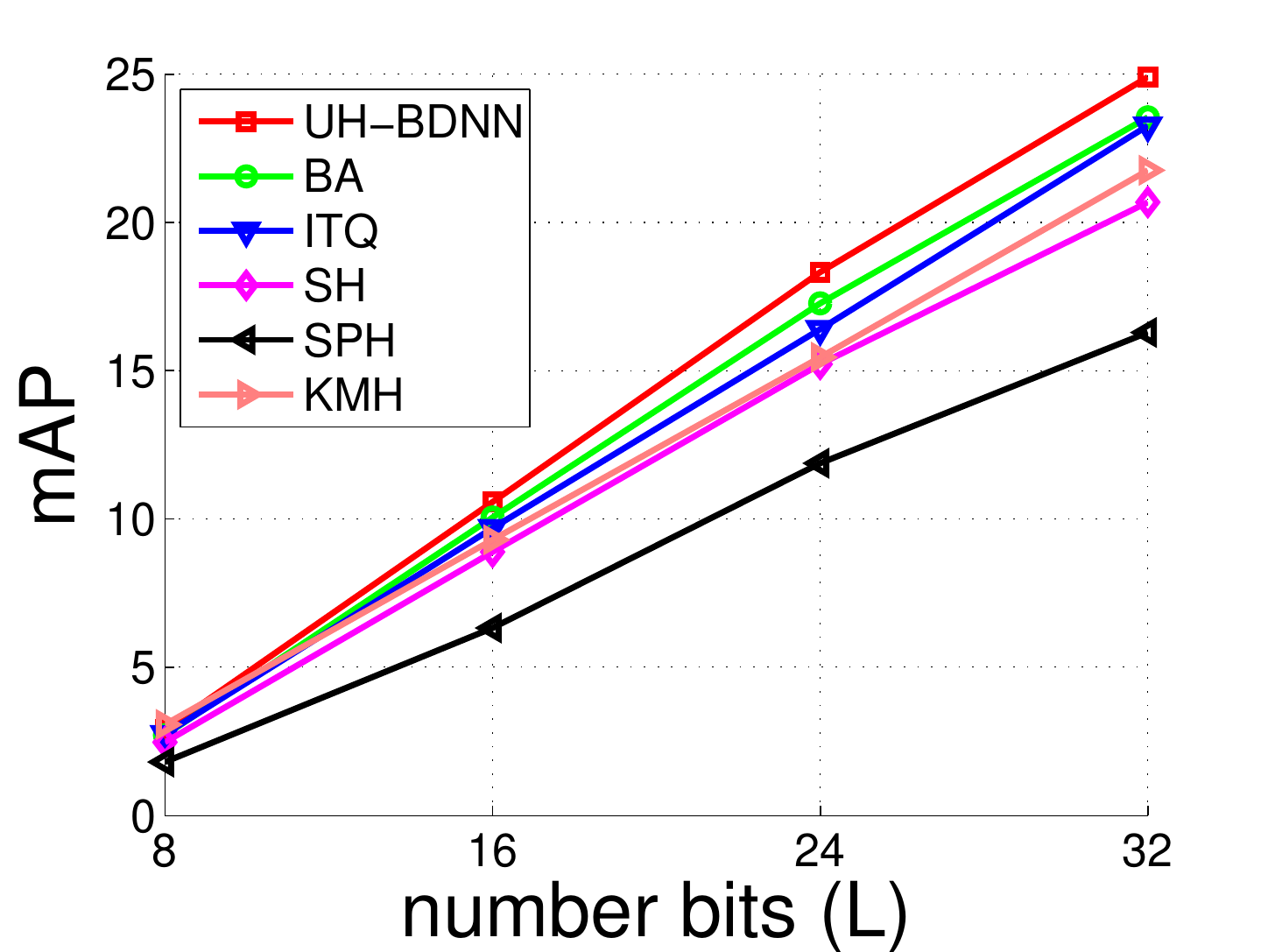} 
\label{fig:mnist_mAP}
}
\subfigure[SIFT1M]{
\includegraphics[scale=0.32]{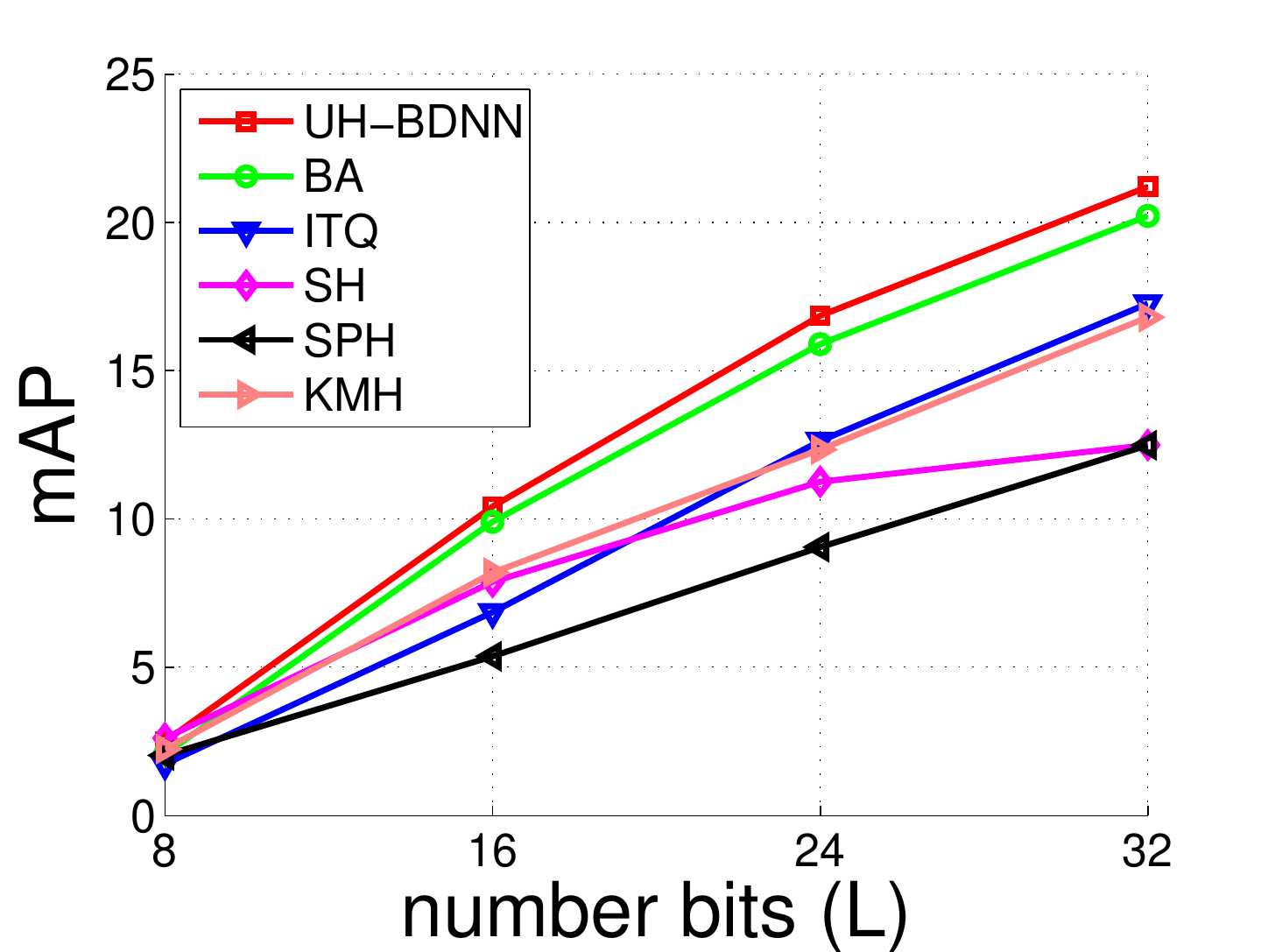} 
\label{fig:sift1m_mAP}
}
\centering
\caption[]{mAP comparison between UH-BDNN and state-of-the-art unsupervised hashing methods on CIFAR10, MNIST, and SIFT1M.}
\label{fig:mAP_cifar10_mnist_sift1m}
\end{figure*}

\begin{table*}[!t]
  \centering
  \caption{Precision at Hamming distance $r=2$ comparison between UH-BDNN and state-of-the-art unsupervised hashing methods on CIFAR10, MNIST, and SIFT1M.}
    \begin{tabular}{|l|c c c c|c c c c|c c c c|}
		\hline
	  \multirow{2}{*}{} & \multicolumn{4}{|c|}{CIFAR10} & \multicolumn{4}{c|}{MNIST} & \multicolumn{4}{c|}{SIFT1M}\\
\cline{1-13}	$L$    &8 &16 &24 &32    &8 &16 &24 &32  &8 &16 &24 &32   \\ \hline 
UH-BDNN	   						 &0.55 &5.79 &22.14 &18.35   &0.53 &6.80 &29.38 &38.50 
    &4.80 &25.20 &62.20 &80.55 \\ \hline
BA \cite{BA_CVPR15}   			 &0.55 &5.65 &20.23 &17.00   &0.51 &6.44 &27.65 &35.29
	&3.85 &23.19 &61.35 &77.15 \\ \hline
ITQ \cite{ITQ} &0.54 &5.05 &18.82 &17.76   &0.51 &5.87 &23.92 &36.35
	&3.19 &14.07 &35.80 &58.69 \\ \hline
SH \cite{SpH}&0.39 &4.23 &14.60 &15.22   &0.43 &6.50 &27.08 &36.69
    &4.67 &24.82 &60.25 &72.40 \\ \hline
SPH \cite{SpH_CVPR12}&0.43 &3.45 &13.47 &13.67   &0.44 &5.02 &22.24 &30.80
	&4.25 &20.98 &47.09 &66.42 \\ \hline
KMH \cite{kmeanHash}	 &0.53 &5.49 &19.55 &15.90   &0.50 &6.36 &25.68 &36.24 			
	&3.74 &20.74 &48.86 &76.04 \\ \hline
	  \end{tabular}
	  \label{tab:soa-UNsup-cifar10-mnist-sift1m-pre}
\end{table*}

\paragraph{Comparison with the state of the art}
The comparative performances in terms of mAP and the precision of Hamming radius $2$ (precision$@2$) are shown in Fig.~\ref{fig:mAP_cifar10_mnist_sift1m} and Table~\ref{tab:soa-UNsup-cifar10-mnist-sift1m-pre}, respectively. 
We find the following observations are consistent across all three datasets. In term of mAP, the proposed UH-BDNN is comparable to or outperforms other methods at all code lengths. At high code lengths, i.e., $L=24,32$, we observe clearer improvements. The best competitor for UH-BDNN is Binary Autoencoder (BA)~\cite{BA_CVPR15}.  
In comparison to BA, at high code lengths, UH-BDNN consistently outperforms BA on all datasets. Regarding the precision$@2$, at $L = 8, 16$, UH-BDNN is comparable to other methods, while at $L = 24, 32$, UH-BDNN achieve considerably better performance than other methods. Specifically, the improvements of UH-BDNN over the best competitor BA~\cite{BA_CVPR15} are clearly observed at $L=32$ on the MNIST and SIFT1M datasets. These comparative results again confirm the advantages of our proposed method, i.e., directly enforcing the independence and balance properties on the binary outputs and carefully relaxing the binary constraint.

\begin{table*}[!t]
\centering
\caption{Comparison with Deep Hashing (DH)~\cite{DeepHash_CVPR15}.} 
\label{tab:compare_DH_cifar512}
\begin{tabular}{|c|c c|c c|c c|c c|}
\hline
&\multicolumn{4}{c|}{CIFAR10} 	 &\multicolumn{4}{c|}{MNIST}\\  \hline
&\multicolumn{2}{c|}{mAP} 	 &\multicolumn{2}{c|}{precision$@2$} &\multicolumn{2}{c|}{mAP} 	 &\multicolumn{2}{c|}{precision$@2$}\\
\cline{2-9}
L	&16	   &32	  &16	 &32    &16    &32    &16    &32 \\\hline
DH~\cite{DeepHash_CVPR15}   &16.17 &16.62 &23.33 &15.77 &43.14 &44.97 &66.10 &73.29\\
\textbf{UH-BDNN} &17.83 &18.52 &24.97 &18.85 &45.38 &47.21 &69.13 &75.26\\\hline 
\end{tabular}
\end{table*} 

\textbf{Comparison with Deep Hashing (DH)~\cite{DeepHash_CVPR15,DeepHash_TIP17}:}
Because the implementation of DH~\cite{DeepHash_CVPR15} has not been released, to make a fair comparison between UH-BDNN and DH, we configure the experiments on CIFAR10 and MNIST similar to~\cite{DeepHash_CVPR15}. Specifically, for each dataset, 1,000 images (i.e., 100 images per class) are randomly sampled as the query set; the remaining images are used as training and database sets. Similar to DH~\cite{DeepHash_CVPR15}, 
GIST descriptors~\cite{gist} are used to represent CIFAR10 images. 
Additionally, the class labels are used as the ground truths for the queries\footnote{It is worth noting that in the evaluation of unsupervised hashing, instead of using the class label as ground truths, most state-of-the-art methods~\cite{ITQ,SpH_CVPR12, kmeanHash, BA_CVPR15} use Euclidean nearest neighbors as ground truths for the queries.}. 
We present the comparative results in term of mAP and precision$@2$ in Table~\ref{tab:compare_DH_cifar512}.
The results show that the proposed UH-BDNN outperforms DH~\cite{DeepHash_CVPR15} at all compared code lengths, in both mAP and precision$@2$.



\textbf{Comparison with DeepBit~\cite{deepbit2016}:}
Here, we compare the proposed UH-BDNN with the recent end-to-end unsupervised hashing DeepBit~\cite{deepbit2016}. 
As reported in~\cite{deepbit2016}, DeepBit uses the pre-trained VGG network~\cite{VGG} and fine-tunes the VGG using 50,000 training samples of CIFAR10. Because DeepBit is an unsupervised method, it does not use the data label during fine-tuning. The comparative results between DeepBit and other methods on the top 1,000 returned images (with the class labels ground truth) on the testing set of CIFAR10 is cited at the top part of Table \ref{tab:deepbit}. 

It is worth noting that in DeepBit~\cite{deepbit2016}, when reporting the results of ITQ, KMH, and SPH, the authors use GIST features for these methods. To make a fair comparison, we evaluate those three hashing methods on the features extracted from the activations of the last fully connected layer of the same pre-trained VGG \cite{VGG} under the same setting. The results of those three methods, which are noted as ITQ-CNN, KMH-CNN, and SPH-CNN, are presented at the bottom part of Table \ref{tab:deepbit}. The results show that using fully-connected features instead of GIST, ITQ-CNN, KMH-CNN, and SPH-CNN provides significant improvements. To evaluate the proposed UH-BDNN, we also use the same fully-connected features. 
The results of UH-BDNN with fully-connected features, which are noted as UH-BDNN-CNN, are presented in the last row of Table \ref{tab:deepbit}. They show that with the same code length, UH-BDNN significantly outperforms the recent end-to-end unsupervised hashing DeepBit \cite{deepbit2016}. 
Furthermore, UH-BDNN also outperforms ITQ-CNN, KMH-CNN, and SPH-CNN with a fair margin. 

\begin{table}[t]
\centering
\footnotesize
\caption{Comparison between DeepBit \cite{deepbit2016} and other unsupervised hashing methods on CIFAR10. The results in the first four rows are cited from \cite{deepbit2016}, which we have also reproduced.}
\begin{tabular}{|c|c|c|c|c|c|} 
\hline
Method    &16 bits &32 bits &64 bits\\ \hline
ITQ \cite{ITQ} &15.67 &16.20  &16.64\\ \hline
KMH \cite{kmeanHash} &13.59 &13.93  &14.46\\ \hline
SPH \cite{SpH_CVPR12} &13.98 &14.58  &15.38\\ \hline
DeepBit \cite{deepbit2016} &19.43 &24.86 &27.73\\ \hline \Xhline{3\arrayrulewidth}
ITQ-CNN &38.52	&41.39 &44.17\\ \hline
KMH-CNN &36.02	&38.18 &40.11\\ \hline
SPH-CNN &30.19	&35.63 &39.23\\ \hline
\textbf{UH-BDNN-CNN} &{40.79} &{44.63} &{46.75}\\ \hline 
\end{tabular}
\label{tab:deepbit}
\end{table}

\section{Supervised Hashing with Binary Deep Neural Network (SH-BDNN)}
\label{sec:SH-BDNN}
One advantage of the proposed UH-BDNN is its flexibility. It can be extended to the supervised version when the label information for the data is available. In this section, to enhance the discriminative power of the binary codes, we extend UH-BDNN to supervised hashing by leveraging the label information. 

To exploit the label information, we follow the approach proposed in Kernel-based Supervised Hashing (KSH)~\cite{SupHashKernel}. The advantage of this approach is  
that it directly encourages the Hamming distances between binary codes of within-class samples equal to $0$, and the Hamming distances between binary codes of between-class samples equal to $L$. 
To achieve this goal, it enforces the 
Hamming distances between learned binary codes to be highly correlated with the pre-computed pairwise label matrix.


Generally, the network structure of SH-BDNN is similar to that of UH-BDNN, excluding the removal of the last layer preserving the reconstruction of UH-BDNN. The layer $n-1$ in UH-BDNN becomes the last layer in SH-BDNN. All desirable properties, i.e., semantic similarity preservation, independence, and balance, in SH-BDNN are constrained on the outputs of its last layer.

\subsection{Formulation of SH-BDNN}
We define the pairwise label matrix $\S$ as
\small
\begin{equation}
\S_{ij} = \left\{ \begin{array}{ll}
1 & \textrm{if $\x_i$ and $\x_j$ are same class}\\
-1 & \textrm{if $\x_i$ and $\x_j$ are not same class}
\end{array} \right.
\label{eq:S}
\end{equation}
\normalsize
To achieve the semantic similarity preserving property, we learn the binary codes such that the Hamming distance between learned codes highly correlates with the matrix $\S$, i.e.,  we want to minimize the quantity  $\norm{\frac{1}{L} (\H^{(n)})^T\H^{(n)} - \S}^2$.
In addition, to achieve the independence and balance properties of codes, we want to minimize the quantities $\norm{\frac{1}{m}\H^{(n)}(\H^{(n)})^T-\I}^2$ and $\norm{\H^{(n)}\1_{m\times 1}}^2$, respectively.

Follow the same reformulation and relaxation as UH-BDNN (Sec.~\ref{subsec:formular_un}), we solve the following constrained optimization which ensures the binary constraint, the semantic similarity preserving, the independence, and the balance properties of codes. 

\vspace{-0.5em}
\footnotesize
\begin{eqnarray}
\min_{\W,\cc,\B} J &=& \frac{1}{2m}\norm{\frac{1}{L} (\H^{(n)})^T\H^{(n)} - \S}^2 +\frac{\lambda_1}{2}\sum_{l=1}^{n-1} \norm{\W^{(l)}}^2  \nonumber \\
{}&&+ \frac{\lambda_2}{2m}\norm{\H^{(n)}-\B}^2+\frac{\lambda_3}{2}\norm{\frac{1}{m}\H^{(n)}(\H^{(n)})^T-\I}^2 \nonumber \\
{}&& +\frac{\lambda_4}{2m}\norm{\H^{(n)}\1_{m\times 1}}^2
 \label{eq:obj_sup2}
\end{eqnarray}
\begin{equation}
\hspace{-1.5cm}\textrm{s.t. }\B \in \{-1,1\}^{L\times m} \label{eq:binary_H1_sup2}
\end{equation}
\normalsize
(\ref{eq:obj_sup2}) under constraint (\ref{eq:binary_H1_sup2}) is our formulation for supervised hashing. The main difference in formulation between UH-BDNN~(\ref{eq:obj_5}) and  SH-BDNN~(\ref{eq:obj_sup2}) is that the reconstruction term preserving the neighbor similarity in UH-BDNN~(\ref{eq:obj_5}) is replaced by the term preserving the label similarity in SH-BDNN~(\ref{eq:obj_sup2}).

\subsection{Optimization}

\begin{algorithm}[!t]
\footnotesize
\caption{Supervised Hashing with Binary Deep Neural Network (SH-BDNN)}
\begin{algorithmic}[1] 
\Require 
\Statex $\X = \{\x_i\}_{i=1}^{m} \in \R^{D\times m}$: labeled training data; 
$L$: code length; $T$: maximum iteration number; $n$: number of layers; $\{s_l\}_{l=2}^{n}$: number of units of layers $2 \to n$ (note: $s_n = L$); $\lambda_1, \lambda_2, \lambda_3, \lambda_4$.
\Ensure 
\Statex 
Parameters $\{\W^{(l)},\cc^{(l)}\}_{l=1}^{n-1}$
\Statex 
\State Compute pairwise label matrix $\S$ using~(\ref{eq:S}).
\State Initialize $\B_{(0)} \in \{-1,1\}^{L\times m}$ using ITQ~\cite{ITQ}
\State Initialize $\{\cc^{(l)}\}_{l=1}^{n-1} = \mathbf{0}_{s_{l+1}\times 1}$. Initialize $\{\W^{(l)}\}_{l=1}^{n-1}$ by getting the top $s_{l+1}$ eigenvectors from the covariance matrix of $\H^{(l)}$. 
\State Fix $\B_{(0)}$, compute $(\W,\cc)_{(0)}$ with $(\W,\cc)$ step using initialized $\{\W^{(l)},\cc^{(l)}\}_{l=1}^{n-1}$ (line 3) as starting point for L-BFGS.
\For{$t = 1 \to T$}
\State Fix $(\W,\cc)_{(t-1)}$, compute $\B_{(t)}$ with $\B$ step
\State Fix $\B_{(t)}$, compute $(\W,\cc)_{(t)}$ with $(\W, \cc)$ step using $(\W,\cc)_{(t-1)}$ as starting point for L-BFGS.
\EndFor
\State Return 
$(\W,\cc)_{(T)}$
\end{algorithmic}
\label{alg2}
\end{algorithm}


To solve~(\ref{eq:obj_sup2}) under constraint~(\ref{eq:binary_H1_sup2}), we use alternating optimization, which comprises two steps over $(\W,\cc)$ and $\B$.

\subsubsection{$(\W,\cc)$ step}
\label{subsub:W_step_sup}
When fixing $\B$, (\ref{eq:obj_sup2}) becomes unconstrained optimization. We used \textit{L-BFGS}~\cite{L-BFGS} optimizer with backpropagation for solving. The gradients of the objective function $J$ (\ref{eq:obj_sup2}) w.r.t. different parameters are computed as follows.

Let us define

\vspace{-0.3em}
\footnotesize
\begin{eqnarray}
\Delta^{(n)} &=& \left[ \frac{1}{mL}\H^{(n)}\left( \V+\V^T \right)+\frac{\lambda_2}{m}\left( \H^{(n)}-\B \right) \right. \nonumber \\
{}&& \left. +\frac{2\lambda_3}{m}\left( \frac{1}{m}\H^{(n)}(\H^{(n)})^T - \I\right)\H^{(n)} \right. \nonumber \\
 {}&& \left. +\frac{\lambda_4}{m}\left( \H^{(n)}\1_{m\times m} \right) \right]\odot f^{(n)'}(\Z^{(n)})
\end{eqnarray}
\normalsize
where \small$\V = \frac{1}{L}(\H^{(n)})^T\H^{(n)} - \S$.\normalsize

\vspace{-0.3em}
\footnotesize
\begin{equation}
\Delta^{(l)} = \left( (\W^{(l)})^T\Delta^{(l+1)} \right) \odot f^{(l)'}(\Z^{(l)}),\forall l = n-1,\cdots,2
\end{equation}
\normalsize
where \small$\Z^{(l)} = \W^{(l-1)}\H^{(l-1)} + \cc^{(l-1)} \1_{1\times m}$, $l=2,\cdots,n$; $\odot$ denotes the Hadamard product. \normalsize

Then $\forall l = n-1,\cdots,1$, we have

\vspace{-0.3em}
\footnotesize
\begin{equation}
\frac{\partial J}{\partial \W^{(l)}} = \Delta^{(l+1)}(\H^{(l)})^T +\lambda_1\W^{(l)}
\end{equation}
\begin{equation}
\frac{\partial J}{\partial \cc^{(l)}} = \Delta^{(l+1)}\1_{m\times 1}
\end{equation}
\normalsize

\subsubsection{$\B$ step}
\label{subsub:B_step_sup}
When fixing $(\W,\cc)$, we can rewrite problem~(\ref{eq:obj_sup2}) as 

\vspace{-0.3em}
\footnotesize
\begin{equation}
\min_{\B} J =  \norm{\H^{(n)}-\B}^2
\label{eq:obj_sup3}
\end{equation}
\vspace{-0.2cm}
\begin{equation}
\textrm{s.t.\ } \B \in \{-1,1\}^{L\times m} \label{eq:binary_H3}
\end{equation}
\normalsize

It is easy to see that the optimal solution for~(\ref{eq:obj_sup3}) under constraint~(\ref{eq:binary_H3}) is $\B = sign(\H^{(n)})$.

The proposed SH-BDNN method is summarized in Algorithm $\mathbf{2}$. In Algorithm $\mathbf{2}$, $\B_{(t)}$ and $(\W,\cc)_{(t)}$ are values of $\B$ and $\{\W^{(l)},\cc^{(l)}\}_{l=1}^{n-1}$ at iteration $t$, respectively.
\begin{figure*}[!t]
\centering
\subfigure[CIFAR10]{
\includegraphics[scale=0.32]{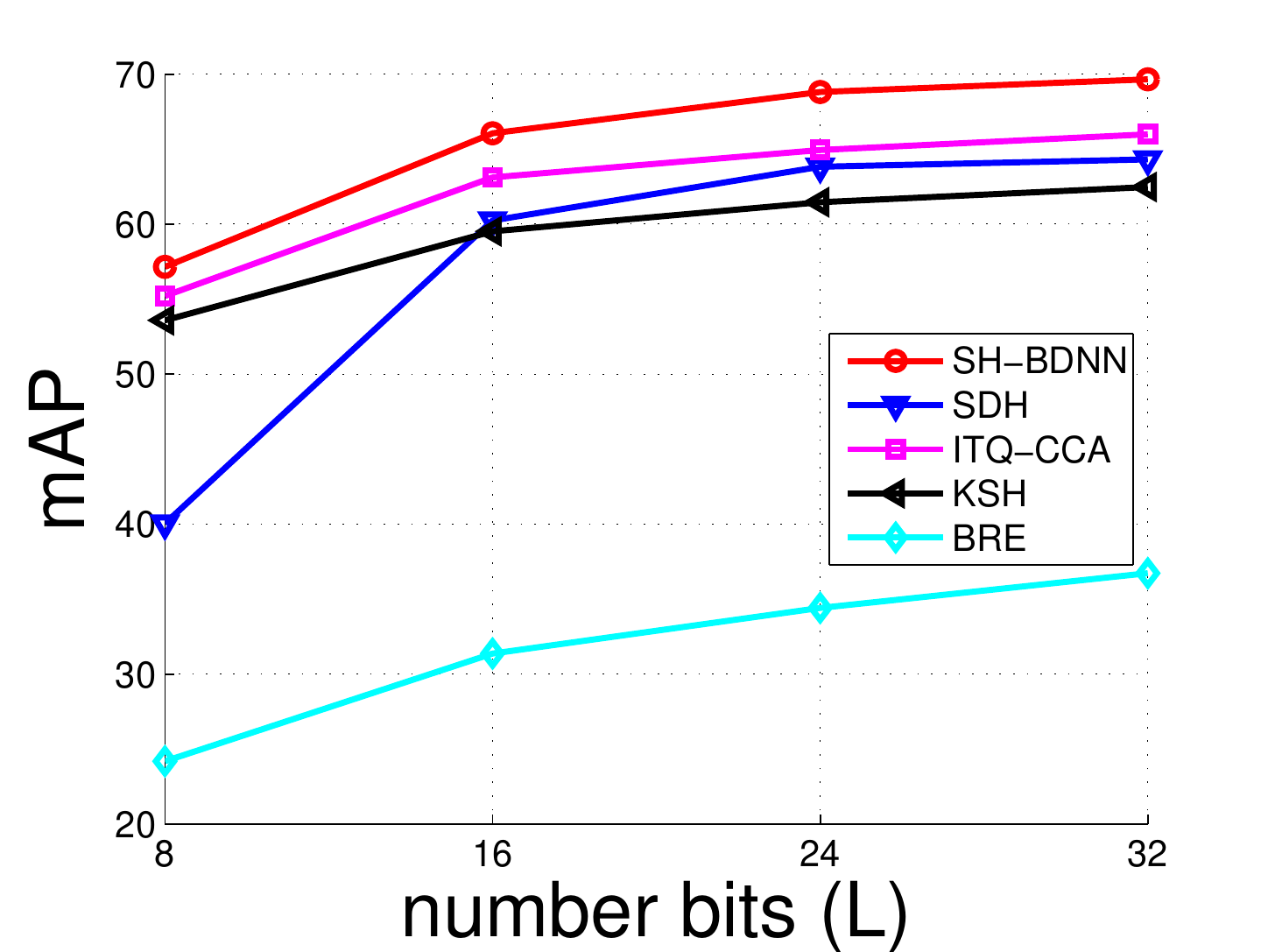}
\label{fig:cifar_mAP_sup}
}
\subfigure[MNIST]{
\includegraphics[scale=0.32]{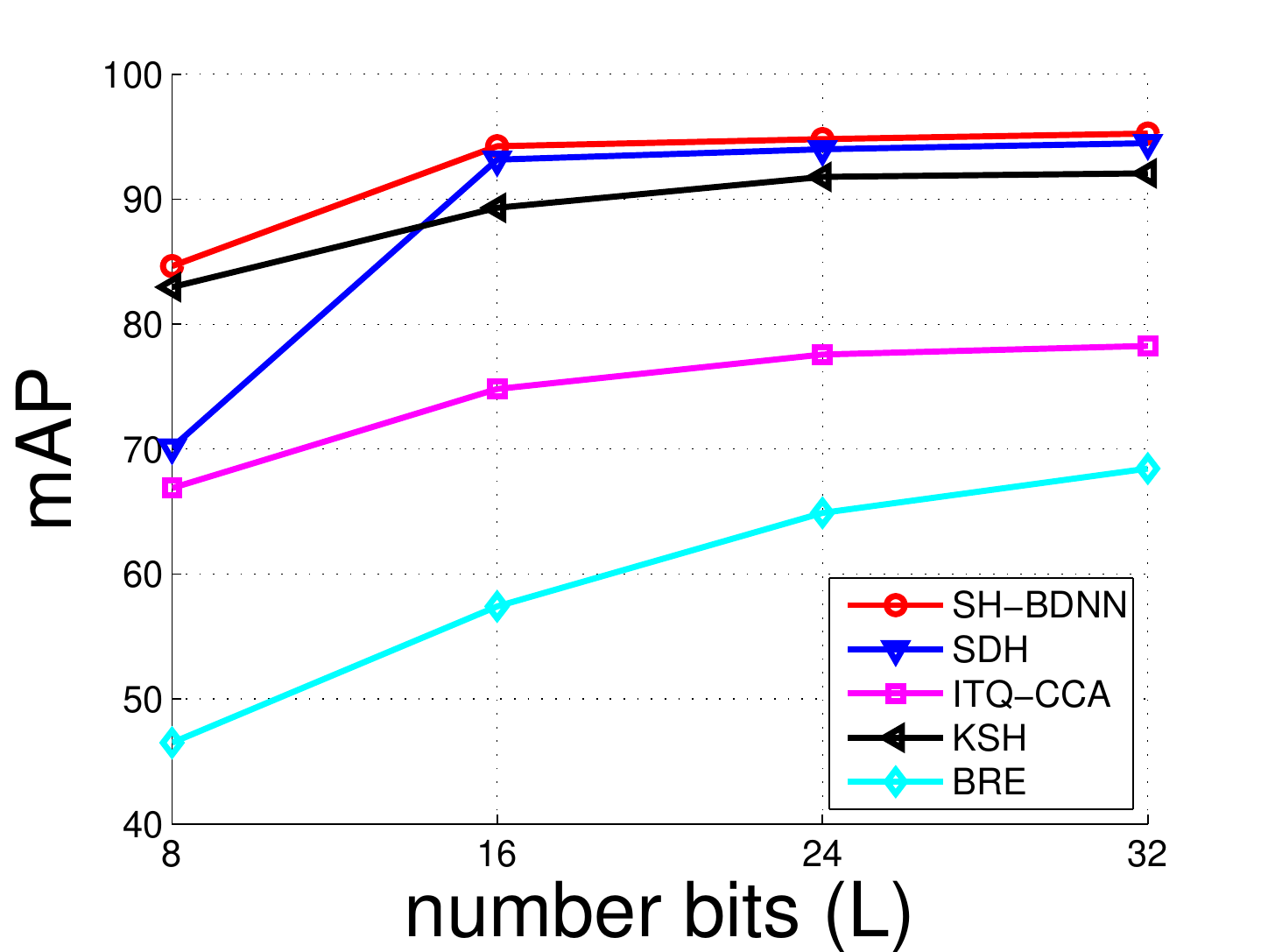} 
\label{fig:mnist_mAP_sup}
}
\subfigure[SUN397]{
\includegraphics[scale=0.32]{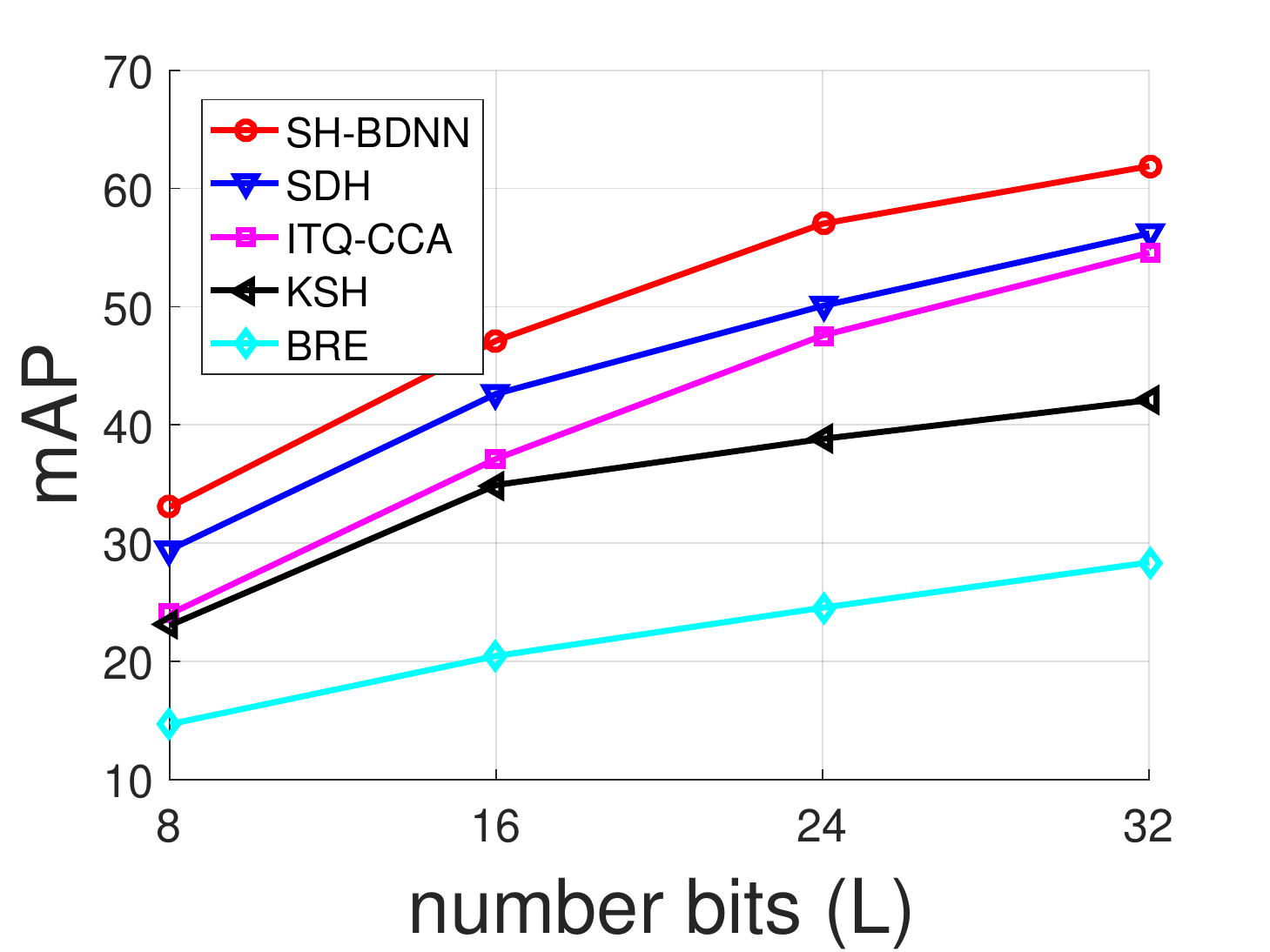} 
\label{fig:sun397_mAP_sup}
}
\caption[]{mAP comparison between SH-BDNN and state-of-the-art supervised hashing methods on CIFAR10, MNIST and SUN397 datasets.}
\label{fig:cifar10_mnist_sup}
\end{figure*}

\begin{table*}[!t]
\centering
\caption{Precision at Hamming distance $r=2$ comparison between SH-BDNN and state-of-the-art supervised hashing methods on CIFAR10, MNIST, and SUN397 datasets.}
\begin{tabular}{|l|c c c c|c c c c|c c c c|}
\hline
\multirow{2}{*}{} & \multicolumn{4}{c|}{CIFAR10} & \multicolumn{4}{c|}{MNIST} & \multicolumn{4}{c|}{SUN397}\\
\cline{1-13}	$L$    		          &8 &16 &24 &32    		    &8 &16 &24 &32     &8 &16 &24 &32  \\ \hline

\textbf{SH-BDNN} & 54.12 & 67.32 & 69.36 & 69.62 & 84.26 & 94.67 & 94.69 & 95.51  
&15.52	&41.98	&52.53	&56.82 \\ \hline
SDH\cite{SDH_CVPR15} & 31.60 & 62.23 & 67.65 & 67.63 & 36.49 & 93.00 & 93.98 & 94.43 
&13.89	&40.39	&49.54	&53.25 \\ \hline  
ITQ-CCA\cite{ITQ}&49.14 &65.68 &67.47 &67.19  &54.35 &79.99 &84.12 &84.57 
&13.22	&37.53	&50.07	&53.12 \\ \hline		                                  
KSH\cite{SupHashKernel}	& 44.81 & 64.08 & 67.01 & 65.76 & 68.07 & 90.79 & 92.86 & 92.41 
&12.64	&40.67	&49.29	&46.45 \\ \hline
BRE\cite{BRE} & 23.84 & 41.11 &47.98 &44.89   &37.67  &69.80 &83.24 &84.61
&9.26	&26.95	&38.36 &40.36 \\ \hline
\end{tabular}
\label{tab:soa-sup-cifar10-mnist-pre}
\end{table*}

\subsection{Evaluation of Supervised Hashing with Binary Deep Neural Network}
\label{sec:eva_SH-BDNN}
In this section, we evaluate and compare our proposed SH-BDNN to the state-of-the-art supervised hashing methods including ITQ-CCA~\cite{ITQ}, Kernel-based Supervised Hashing (KSH)~\cite{SupHashKernel}, Binary Reconstructive Embedding (BRE)~\cite{BRE}, and Supervised Discrete Hashing (SDH)~\cite{SDH_CVPR15}. We use the released implementations and the suggested parameters provided by the authors for all compared methods.

\subsubsection{Dataset, evaluation protocol, and implementation notes}
\paragraph{Dataset} We evaluate and compare methods on the CIFAR-10, MNIST, and SUN397 datasets. The descriptions of the first two datasets are presented in section~\ref{subsec:data-imp-eva}. 

The SUN397~\cite{sun397} dataset contains approximately 108,000 images from 397 scene categories. We select 42 categories that contain more than 500 images, which results in a dataset of approximately 35,000 images in total. We then randomly sample 100 images per class from the dataset to form a query set of 4,200 images. The remaining images are used as the training set and the database set. 
Each image is represented by an 800-dimensional feature vector extracted by PCA from 4096-dimensional CNN feature produced by AlexNet~\cite{DBLP:conf/nips/KrizhevskySH12}. 

\paragraph{Evaluation protocol} We report the retrieval performances by using the two standard metrics in the literature~\cite{SDH_CVPR15, ITQ,SupHashKernel}: precision of Hamming radius $2$ (precision$@2$) and mean Average Precision (mAP).

\paragraph{Implementation notes} The network configuration of SH-BDNN is similar to the configuration of UH-BDNN except the final layer is removed. The parameters $\lambda_1$, $\lambda_2$, $\lambda_3$, and $\lambda_4$ are empirically set using cross validation as $10^{-3}$, $5$, $1$, and $10^{-4}$, respectively. The max iteration number $T$ is empirically set to 5. 

Following the settings in ITQ-CCA~\cite{ITQ} and SDH~\cite{SDH_CVPR15},  all training samples are used in the learning for these two methods. For SH-BDNN, KSH~\cite{SupHashKernel} and BRE~\cite{BRE} where the label information is leveraged by the pairwise label matrix, we randomly select 300 training samples from each class to form a training set. In the supervised setting, the class label is used to define the ground truths of queries.

\subsubsection{Retrieval results}
Fig.~\ref{fig:cifar10_mnist_sup} and Table~\ref{tab:soa-sup-cifar10-mnist-pre} show comparative results between the proposed SH-BDNN and other supervised hashing methods on the CIFAR10, MNIST, and SUN397 datasets. 

On the CIFAR10 dataset, Fig.~\ref{fig:cifar_mAP_sup} and Table~\ref{tab:soa-sup-cifar10-mnist-pre} show that our proposed SH-BDNN clearly outperforms all compared methods at all code lengths by a fair margin in both evaluation metrics, i.e., mAP and precision@2. The best competitor to SH-BDNN on this dataset is CCA-ITQ~\cite{ITQ}. The more improvements of SH-BDNN over CCA-ITQ are observed at high code lengths, i.e., SH-BDNN outperforms CCA-ITQ by approximately 4\% at $L=24$ and $L=32$. 


On the MNIST dataset, Fig.~\ref{fig:mnist_mAP_sup} and Table~\ref{tab:soa-sup-cifar10-mnist-pre} show that the proposed SH-BDNN significant outperforms the current state-of-the-art SDH~\cite{SDH_CVPR15} at low code lengths, i.e., $L=8$. At higher code lengths, however, the performances of SH-BDNN and SDH~\cite{SDH_CVPR15} are comparable. Moreover, SH-BDNN significantly outperforms other methods, i.e., ITQ-CCA~\cite{ITQ},  KSH~\cite{SupHashKernel}, and BRE~\cite{BRE}, on both mAP and precision@2.

On the SUN397 dataset, the proposed SH-BDNN outperforms other competitors at all code lengths in terms of both mAP and precision@2. The best competitor to SH-BDNN on this dataset is SDH~\cite{SDH_CVPR15}. At high code lengths (e.g., $L=24, 32$), SH-BDNN achieves more improvements over SDH.

\section{Supervised Hashing with End-to-End Binary Deep Neural Network (E2E-BDNN)}
\label{sec:SH-E2E-BDNN}


Even though the proposed  SH-BDNN can significantly enhance the discriminative power of the binary codes, similar to other hashing methods, its capability is partially dependent on the discriminative power of the image features. The recent end-to-end deep learning-based supervised hashing methods~\cite{DSRH,DRSCH,simulfeature} have shown that joint learning image representations and binary hash codes in an end-to-end fashion can boost the retrieval accuracy. 
Therefore, in this section, we propose to extend the proposed SH-BDNN to an end-to-end framework. 
Specifically, we integrate the convolutional neural network (CNN) with our supervised hashing network (SH-BDNN) into an unified end-to-end deep architecture, namely, the End-to-End Binary Deep Neural Network (E2E-BDNN), which can jointly learn visual features and binary representations of images. 
In the following, we first introduce our proposed network architecture. We then describe the training process. Finally, we present   experiments on various benchmark datasets.

\subsection{Network architecture}
\label{ssec:netarch}

The network consists of three main components: \textit{(i)} a feature extractor, \textit{(ii)} a dimensional reduction layer, and \textit{(iii)} a binary optimizer component. We utilize the AlexNet model~\cite{DBLP:conf/nips/KrizhevskySH12} as the feature extractor component of the E2E-BDNN.  
In our configuration, we remove the last layer of AlexNet, namely the {\tt softmax} layer, and consider its last fully connected layer ({\tt fc7}) as the image representation.

The dimensional reduction component (the DR layer) involves a fully connected layer for reducing the high dimensional image representations outputted by the feature extractor component into lower dimensional representations. We use the identity function as the activation function for this DR layer. \red{Thus, the DR layer performs a linear  projection to reduce the  dimension of AlexNet features.
}
The reduced representations are then used as inputs for the following binary optimizer component. 

%
%
%

The binary optimizer component of E2E-BDNN is similar to SH-BDNN. Thus, we also constrain the output codes of E2E-BDNN to be binary. These codes also have the desired properties such as semantic similarity preservation, independence, and balance. 
By using the same design as SH-BDNN for the last component of E2E-BDNN, it allows us to observe the advantages of the end-to-end architecture over SH-BDNN.

The training data for the E2E-BDNN are labelled images, contrasting with SH-BDNN which uses visual features such as GIST~\cite{gist}, SIFT~\cite{SIFT_Lowe} or deep features from  convolutional deep networks. 
Given the input labeled images, we aim to learn binary codes with the aforementioned desired properties, i.e., semantic similarity preservation, independence, and balance. To achieve these properties on the codes, we use a similar objective function as SH-BDNN. However, it is important to mention that in SH-BDNN, by its non end-to-end architecture, we can feed the whole training set into the network at a time during training, which does not hold for E2E-BDNN. Due to the memory consumption of the end-to-end architecture, we can only feed a minibatch of images into the network at a time during training. Technically, let $\H$ be the output of the last fully connected layer of E2E-BDNN for a minibatch of size $m_b$; $\S$ be the similarity matrix defined over the minibatch (using equation~(\ref{eq:S})); and $\B$ serve as an auxiliary variable. Similar to SH-BDNN, we train the network to minimize the following constrained loss function

\vspace{-0.3em}
\footnotesize
\begin{eqnarray}
\min_{\W ,\B} J &=& \frac{\lambda_1}{2m_b}\norm{\frac{1}{L} \H^T\H - \S}^2 \nonumber \\
{}&&+ \frac{\lambda_2}{2m_b}\norm{\H-\B}^2+\frac{\lambda_3}{2}\norm{\frac{1}{m_b}\H\H^T-\I}^2 \nonumber \\
{}&& +\frac{\lambda_4}{2m_b}\norm{\H\1_{m_b\times 1}}^2
 \label{eq:obj_supe2e}
\end{eqnarray}
\begin{equation}
\hspace{-1.5cm}\textrm{s.t. }\B \in \{-1,1\}^{L\times m_b} \label{eq:binary_H1_supe2e}
\end{equation}
\normalsize
 

\subsection{Training}
\label{sec:e2etraining}

\begin{algorithm}[!t]
	\footnotesize
	\caption{End-to-End Binary Deep Neural Network (E2E-BDNN) Learning}
	\begin{algorithmic}[1]
		\Require
			\Statex $\X =\{\x_i\}_{i=1}^m$: labeled training images; $m_b$: minibatch size; 
            $L$: code length;
            $K$, $T$: maximum iteration.
            $\lambda_1, \lambda_2, \lambda_3, \lambda_4$: hyperparameters.
		\Ensure
			\Statex
                Network parameters $\W$
			\Statex
            \State Initialize the network $\W_{(0)}^{(0)}$
            \State Initialize $\B^{(0)} \in \{-1,1\}^{L\times m}$ via ITQ~\cite{ITQ}
			\For{$k = 1 \to K$}
                \For {$t = 1 \to T $}
                    \State A minibatch $\X_{(t)}$ is sampled from $\X$
                    \State Compute the corresponding similarity matrix $\S_{(t)}$
                    \State From $\B^{(k-1)}$, sample $\B_{(t)}$ corresponding to $\X_{(t)}$
                    \State Fix $\B_{(t)}$, optimize $\W_{(t)}^{(k)}$ via SGD
                \EndFor
                \State Update $\B^{(k)}$ by $\W_{(T)}^{(k)}$
            \EndFor

			\State Return $\W_{(T)}^{(K)}$
    \end{algorithmic}
\label{e2ealg}
\end{algorithm}


The training procedure for E2E-BDNN is presented in Algorithm $\mathbf{3}$. In Algorithm $\mathbf{3}$, $\X_{(t)}$ and $\B_{(t)} \in \{-1, 1\}^{L \times m_b}$ are a minibatch sampled from the training set at iteration $t$ and its corresponding binary codes, respectively. $\B^{(k)}$ is the binary codes of the whole training set $\X$ at iteration $k$; and $\W_{(t)}^{(k)}$ is the network weight when learning up to iterations $t$ and $k$. 

At first (line 1 in Algorithm $\mathbf{3}$), we initialize the network parameter $\W_{(0)}^{(0)}$ as follows. \textit{(i)} The feature extractor component is initialized by the pretrained AlexNet model~\cite{DBLP:conf/nips/KrizhevskySH12}. 
\textcolor{black}{\textit{(ii)} The dimensional reduction (DR) layer is initialized by the top eigenvectors extracted from the covariance matrix of the AlexNet features (i.e., the outputs of {\tt fc7} layer) of the training set.}
\textit{(iii)} \redd{The binary optimizer component is initialized by the trained SH-BDNN models (in Section \ref{sec:SH-BDNN})}.

We then initialize the binary code matrix of the whole dataset $\B^{(0)} \in \{-1,1\}^{L\times n}$ via ITQ~\cite{ITQ} (line 2 in the Algorithm $\mathbf{3}$). Here, AlexNet features are used as training inputs for ITQ. 



In each iteration $t$ of Algorithm $\mathbf{3}$, we only sample a minibatch $\X_{(t)}$ from the training set to feed into the network (line 5 in Algorithm $\mathbf{3}$). Thus, after $T$ iterations, we exhaustively sample all training data. In each iteration $t$, we first create the similarity matrix $\S_{(t)}$ (using equation~(\ref{eq:S})) corresponding to $\X_{(t)}$, as well as the $\B_{(t)}$ matrix (lines 6 and 7 in Algorithm $\mathbf{3}$). Since $\B_{(t)}$ has already been computed, we can fix that variable and optimize the network parameter $\W_{(t)}^{(k)}$ by standard backpropagation with Stochastic Gradient Descent (SGD) (line 8 in Algorithm $\mathbf{3}$).
After $T$ iterations, since the network was exhaustively learned from the whole training set, we update $\B^{(k)} = sign(\F)$ (line 10 in the Algorithm $\mathbf{3}$), where $\F$ is the outputs of the last fully connected layer for all training samples. We then repeat the optimization procedure until it reaches a criterion, i.e., after $K$ iterations.

\textbf{Implementation details}
The proposed E2E-BDNN is implemented in MATLAB with MatConvNet library \cite{matconvneet}. All experiments are conducted on a workstation machine with a GPU Titan X. Regarding the hyperparameters of the loss function, 
we empirically set $\lambda_1 = 10^{-1}$, $\lambda_2=10^{-2}$, $\lambda_3=10^{-2}$ and $\lambda_4=10^{-3}$. 
The learning rate is set to $10^{-4}$ and the weight decay is set to $5\times 10^{-4}$. The minibatch size is $256$.


\begin{table*}[!t]
\centering
\caption{mAP comparison between SH-BDNN and E2E-BDNN on CIFAR10, MNIST, and SUN397 datasets.}
\begin{tabular}{|l|c c c c|c c c c|c c c c|}
\hline
\multirow{2}{*}{} & \multicolumn{4}{c|}{CIFAR10} & \multicolumn{4}{c|}{MNIST} & \multicolumn{4}{c|}{SUN397}\\
\cline{1-13}	$L$   &8 &16 &24 &32&8 &16 &24 &32     &8 &16 &24 &32  \\ \hline
\textbf{SH-BDNN} &57.15 &66.04 &68.81 &69.66    & 84.65 & 94.24 & 94.80 & 95.25  
&33.06  &47.13  &57.02   &61.89\\ \hline
\textbf{E2E-BDNN} &64.83 &71.02 &72.37 &73.56 & 88.82 & 98.03 & 98.16 & 98.21 
&34.15  &48.21  &59.51   &64.58\\ \hline  
\end{tabular}
\label{tab:map-e2e}
\end{table*}
\subsection{Evaluation of End-to-End Binary Deep Neural Network (E2E-BDNN)}
\label{ssec:e2e_exp}
Since we have already compared SH-DBNN to other supervised hashing methods in Section~\ref{sec:eva_SH-BDNN}, in this experiment we focus on comparing E2E-BDNN with SH-BDNN. We also compare the proposed E2E-BDNN to other end-to-end hashing methods~\cite{DRSCH,DSRH,simulfeature,DQN,DHN,DPSH}.

\paragraph{Comparison between SH-BDNN and E2E-BDNN}
Table~\ref{tab:map-e2e} presents the comparative mAP between SH-BDNN and E2E-BDNN. The results shows that E2E-BDNN consistently improves over SH-BDNN at all code lengths on all
datasets. The large improvements of E2E-BDNN over SH-BDNN are observed on the CIFAR10 and MNIST datasets, especially at low code lengths, i.e., on the CIFAR10 dataset, E2E-BDNN outperforms SH-BDNN by 7.7\% and 5\% at $L=8$ and $L=16$, respectively; on the MNIST dataset, E2E-BDNN outperforms SH-BDNN by 4.2\% and 3.8\% at $L=8$ and $L=16$, respectively. On the SUN397 dataset, the improvements of E2E-BDNN over SH-BDNN are clearer at high code lengths, i.e., E2E-BDNN outperforms SH-BDNN by 2.5\% and 2.7\% at $L=24$ and $L=32$, respectively.
The improvements of E2E-BDNN over SH-BDNN confirm the effectiveness of the proposed end-to-end architecture for learning discriminative binary codes. 

\paragraph{Comparison between E2E-BDNN and other end-to-end supervised hashing methods}

\begin{table}[!t]
\centering
\caption{mAP comparison between E2E-BDNN, SH-BDNN and DNNH\cite{simulfeature}, DQN\cite{DQN}, DPSH \cite{DPSH}, DHN \cite{DHN} on CIFAR10 (Setting 1).}
\begin{tabular}{|c|c c c|}
\hline 
$L$                 &24 &32 &48 \\ \hline
\textbf{E2E-BDNN}  & 60.02 & 61.35 & 63.59 \\ \hline
\textbf{SH-BDNN}   & 57.30 & 58.66 & 60.08 \\ \hline
DNNH\cite{simulfeature} & 56.60 & 55.80 & 58.10 \\ \hline
DQN\cite{DQN} & 55.80 & 56.40 & 58.00 \\ \hline
DPSH \cite{DPSH} & 57.57 & 58.54 & 60.17 \\ \hline
DHN\cite{DHN} & 59.40 & 60.30 & 62.10 \\ \hline
\end{tabular}
\label{tab:e2e_simul}
\end{table}

\begin{table}[!t]
\centering
\caption{mAP comparison between E2E-BDNN, SH-BDNN and DSRH\cite{DSRH}, DRSCH\cite{DRSCH} on CIFAR10 (Setting 2).}
\begin{tabular}{|c|c c c|}
\hline
$L$ & 24 & 32 & 48 \\ \hline
\textbf{E2E-BDNN} & 67.16 & 68.72 & 69.23 \\ \hline
\textbf{SH-BDNN} & 65.21 & 66.22 & 66.53 \\ \hline
DSRH~\cite{DSRH} & 61.08 & 61.74 & 61.77 \\ \hline
DRSCH~\cite{DRSCH}  &62.19 &62.87 &63.05 \\ \hline
\end{tabular}
\label{tab:e2e_bitscale}
\end{table}

We also compare our proposed deep networks SH-BDNN and E2E-BDNN with other end-to-end supervised hashing architectures, i.e., Hashing with Deep Neural Network (DNNH)~\cite{simulfeature}, Deep Hashing Network (DHN)~\cite{DHN}, Deep Quantization Network (DQN)~\cite{DQN}, Deep Semantic Ranking Hashing (DSRH)~\cite{DSRH}, Deep Pairwise Supervised Hashing (DPSH) \cite{DPSH}, and Deep Regularized Similarity Comparison Hashing (DRSCH)~\cite{DRSCH}.
In those works, the authors propose the  frameworks in which the image features and hash codes are simultaneously learned by combining a CNN and a binary quantization layer into a unified model. 
However, their binary mapping layer only applies a simple operation, e.g., an approximation of the $sign$ function ($logistic$~\cite{DSRH,simulfeature}, $tanh$~\cite{DRSCH}), $l_1$ norm approximation of binary constraints~\cite{DHN}. Our SH-BDNN and E2E-BDNN advances over those works in the way to map the image features to the binary codes. Furthermore, our learned codes ensure good properties, i.e. independence and balance, while \cite{simulfeature,DRSCH,DQN,DHN,DPSH} do not consider such properties, and~\cite{DSRH} only considers the balance of codes. 
It is worth noting that, in~\cite{simulfeature,DSRH,DRSCH,DHN,DPSH}, different settings are used for evaluation. For a fair comparison, following those works, we setup two different experimental settings on CIFAR10 as follows

\begin{itemize}
    \item Setting 1: following ~\cite{simulfeature,DQN,DHN}, we randomly sample 100 images per class to form 1K testing images. The remaining 59K images are used as database images. 
Furthermore, 500 images per class are sampled from the database to form 5K training images.
    \item Setting 2: following~\cite{DSRH,DRSCH}, we randomly sample 1K images per class to form 10K testing images. The remaining 50K images serve as the training set. In the test phase, each test image is searched through the test set by the leave-one-out procedure. 
\end{itemize}


Table~\ref{tab:e2e_simul} shows the comparative mAP between our methods and DNNH~\cite{simulfeature}, DQN~\cite{DQN}, DPSH \cite{DPSH}, and DHN \cite{DHN} on the CIFAR10 dataset with setting 1. The results show that even with the non end-to-end approach, our SH-BDNN outperforms DNNH and DQN and is comparable to DPSH at all code lengths.
With the end-to-end approach, it helps to boost the performance of the SH-BDNN. The proposed E2E-BDNN outperforms all compared methods, DNNH~\cite{simulfeature}, DHN~\cite{DHN}, DQN~\cite{DQN}, and DPSH \cite{DPSH}. 
It is worth noting that in \cite{simulfeature}, increasing the code length does not necessarily boost the retrieval accuracy, i.e., \cite{simulfeature} reports a mAP of 55.80 at $L=32$, while a higher mAP, i.e., 56.60 is reported at $L=24$. 
In contrast to~\cite{simulfeature}, both SH-BDNN and E2E-BDNN improve mAP when increasing the code length. 
\red{Additionally, we also observe that both DPSH and DHN maximize log-likelihood objective functions to ensure that (dis)similar input pairs result in (dis)similar output pairs. Hence, the large performance gaps between DHN and DPSH show that the optimization method proposed in DPSH does not well handle the binary constraint. Specifically, DPSH resorts to the $sign$ function to obtain the binary codes during optimization but ignores its ill-posed gradient problem.}
\red{More importantly, the superior performance of our proposed method over the compared methods confirms the effectiveness of the proposed approach in dealing with binary constrains and the provision of desired properties such as independence and balance on the produced codes.}

Table~\ref{tab:e2e_bitscale} presents the comparative mAP between the proposed SH-BDNN, E2E-BDNN and the competitors DSRH~\cite{DSRH}, DRSCH~\cite{DRSCH} on the CIFAR10 dataset with setting 2.
The results clearly show that the proposed E2E-BDNN significantly outperforms DSRH~\cite{DSRH} and DRSCH~\cite{DRSCH} at all code lengths. Compared with the best competitor DRSCH~\cite{DRSCH}, the improvements of E2E-BDNN over DRSCH range from 5\% to 6\% at different code lengths. Furthermore, we can see that even with the non end-to-end approach, the proposed SH-BDNN also outperforms DSRH~\cite{DSRH} and DRSCH~\cite{DRSCH}.

\section{Conclusion}
\label{sec:conclusion}

In this paper, we propose three deep hashing neural networks for learning compact binary presentations. Firstly, we introduce UH-BDNN and SH-BDNN for unsupervised and supervised hashing respectively. In our novel designs, the networks are constrained to directly produce binary codes at one layer. The designs also ensure good properties for produced codes, i.e.,  similarity preservation, independence, and balance. Together with the designs, we also propose alternating optimization schemes that allow us  to effectively deal with binary constraints on the codes.  
We then propose to extend SH-BDNN to an end-to-end deep hashing framework (E2E-BDNN) that jointly learns the image representations and the binary codes. 
The solid experimental results on various benchmark datasets show that the proposed methods compare favorably or outperform state-of-the-art hashing methods. 

\section*{Acknowledgement}
This research was supported by the National Research Foundation Singapore under its AI Singapore Programme (Award number: AISG-100E-2018-005). This work was also supported by both ST Electronics and the National Research Foundation (NRF), Prime Minister's Office, Singapore under Corporate Laboratory at University Scheme (Programme Title: STEE Infosec - SUTD Corporate Laboratory).

\bibliographystyle{IEEEtran}
\bibliography{hash}
\vspace{-1cm}
\begin{IEEEbiography}[{\includegraphics[width=0.9in,height=1.1in,clip,keepaspectratio]{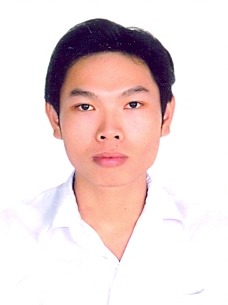}}]
{Thanh-Toan Do}
is  currently  a lecturer at the Department of Computer Science, University of Liverpool (UoL), United Kingdom. He obtained a  Ph.D.  in  Computer  Science  from  INRIA, Rennes, France in 2012. Before joining UoL, he was a Research Fellow at the Singapore University of Technology  and  Design,  Singapore (2013 - 2016) and the University of Adelaide,  Australia (2016 - 2018).  His research interests include  Computer Vision and Machine Learning.
\end{IEEEbiography}

\vspace{-1.5cm}
\begin{IEEEbiography}[{\includegraphics[width=1in,height=1.25in,clip,keepaspectratio]{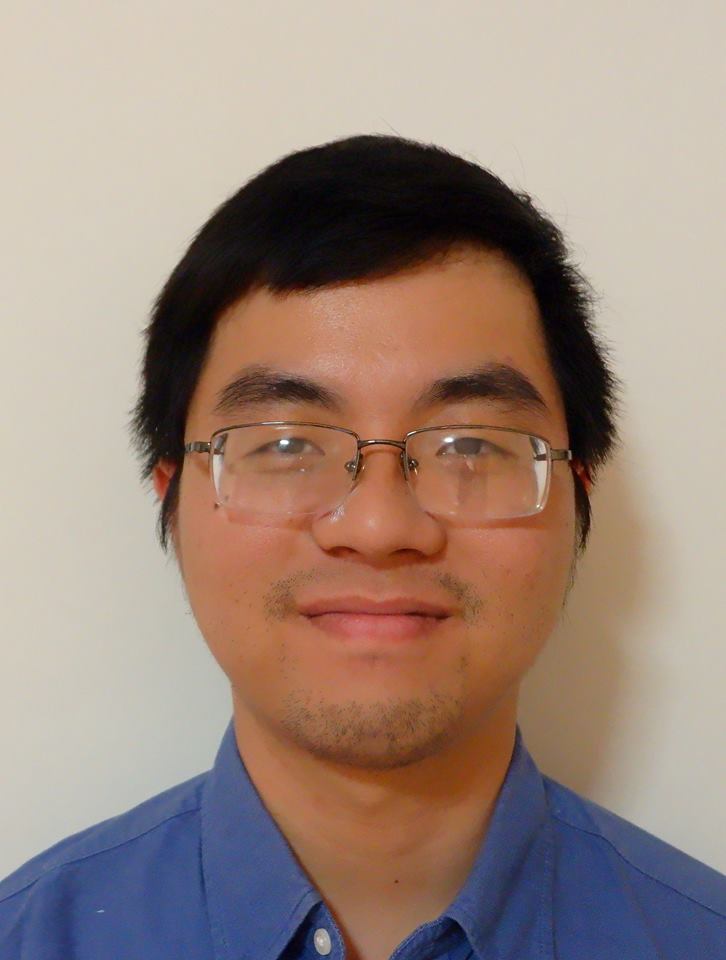}}]
{Tuan Hoang}
is currently a Ph.D. student at Singapore University of Technology and Design (SUTD). Before joining SUTD, he received the bachelor degree in Electrical Engineering from Portland State University, in 2014. His research interests are content-based image retrieval and image hashing.
\end{IEEEbiography}

\vspace{-1cm}
\begin{IEEEbiography}[{\includegraphics[width=1in,height=1.2in,clip,keepaspectratio]{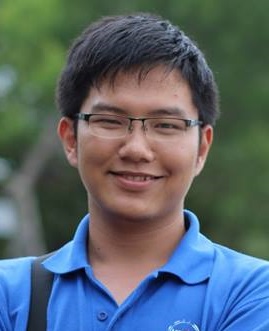}}]
{Khoa Le}
 received a BSc for an honours degree from the University of Science, Vietnam National University, in 2015. Since 2016, he has been a research assistant at Singapore University of Technology and Design (SUTD). His current research interests are deep learning and image retrieval.
\end{IEEEbiography}

\vspace{-1cm}
\begin{IEEEbiography}[{\includegraphics[width=1in,height=1.26in,clip,keepaspectratio]{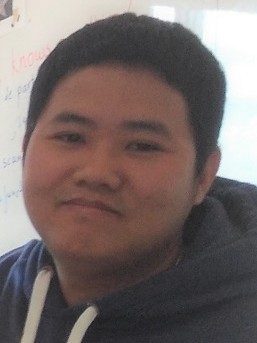}}]
{Anh-Dzung Doan} is currently a Ph.D. student at the University of Adelaide (UoA). He received a BSc for an honours degree from the University of Science, Vietnam National University, in 2013. Before joining UoA, he was a research assistant at Singapore University of Technology and Design (SUTD) (2014-2017). His research interests include 3D computer vision, robotic vision, and image retrieval.
\end{IEEEbiography}

\vspace{-1.5cm}
\begin{IEEEbiography}[{\includegraphics[width=1in,height=1.25in,clip,keepaspectratio]{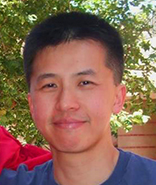}}]
{Ngai-Man Cheung}
received a Ph.D. degree in electrical engineering from the University  of Southern California, Los Angeles, CA, in 2008. He is currently an Associate Professor with the Singapore University of Technology and Design (SUTD). From 2009 - 2011, he was a postdoctoral researcher with the Image, Video and Multimedia Systems group at Stanford University, Stanford, CA.  He has also held research positions with Texas Instruments Research Center Japan, Nokia Research Center, IBM T. J. Watson Research Center, HP Labs Japan, Hong Kong University of Science and Technology (HKUST), and Mitsubishi Electric Research Labs (MERL). His work has resulted in 10 U.S. patents granted with several pending. His research interests include signal, image, and video processing, and computer vision.
\end{IEEEbiography}

\end{document}